\documentclass[10pt,journal,compsoc]{IEEEtran}

\ifCLASSOPTIONcompsoc
  \usepackage[nocompress]{cite}
\else
  \usepackage{cite}
\fi
\usepackage{amsmath}
\usepackage{acronym}
\usepackage{algorithmic}

\usepackage{array}

\usepackage{mdwmath}
\usepackage{mdwtab}

\usepackage{eqparbox}

\ifCLASSOPTIONcompsoc
 \usepackage[caption=false,font=footnotesize,labelfont=sf,textfont=sf]{subfig}
\else
 \usepackage[caption=false,font=footnotesize]{subfig}
\fi
\usepackage{fixltx2e}

\usepackage{stfloats}
\ifCLASSOPTIONcaptionsoff
 \usepackage[nomarkers]{endfloat}
\let\MYoriglatexcaption\caption
\renewcommand{\caption}[2][\relax]{\MYoriglatexcaption[#2]{#2}}
\fi
\usepackage{url}

\usepackage{amsmath,amsfonts}
\usepackage{array}
\usepackage{booktabs}
\usepackage{textcomp}
\usepackage{url}
\usepackage{verbatim}
\usepackage{amsthm}
\usepackage{url}
\usepackage{graphicx}
\usepackage{multirow}
\usepackage{mathtools}
\usepackage{todonotes}
\usepackage{enumitem}
\usepackage{xcolor}
\usepackage{hyperref}
\usepackage[capitalize,noabbrev]{cleveref}
\usepackage{xspace}
\usepackage{etoc}
\usepackage{tikz}
\usepackage[]{mdframed}

\theoremstyle{plain}
\newtheorem{theorem}{Theorem}[section]

\theoremstyle{definition}
\newtheorem{definition}[theorem]{Definition}

\theoremstyle{remark}

\newcommand{\method}{NMTune\xspace}

\newcommand{\revision}[1]{{\color{black}{#1}}}

\begin{document}
%
\title{\revision{Impact of Noisy Supervision in Foundation Model Learning}}
%
%
%
%

\author{Hao Chen,  Zihan Wang, Ran Tao, Hongxin Wei, 
Xing Xie,~\IEEEmembership{Fellow,~IEEE,} \\  Masashi Sugiyama,~\IEEEmembership{Senior Member,~IEEE,} Bhiksha Raj,~\IEEEmembership{Fellow,~IEEE,} Jindong Wang,~\IEEEmembership{Member,~IEEE}
\IEEEcompsocitemizethanks{
\IEEEcompsocthanksitem Hao Chen, Zihan Wang, Ran Tao, and Bhiksha Raj are with Carnegie Mellon University. E-mail: haoc3@andrew.cmu.edu, zihanwa3@cs.andrew.cmu, taoran1@cmu.edu, bhikshar@andrew.cmu.edu.
\IEEEcompsocthanksitem Xing Xie is with Microsoft Research Asia. Email: xingx@microsoft.com
\IEEEcompsocthanksitem Hongxin Wei is with Southern University of Science and Technology. Email: weihx@sustech.edu.cn.
\IEEEcompsocthanksitem Masashi Sugiyama is with RIKEN and the University of Tokyo. Email: sugi@k.u-tokyo.ac.jp.
\IEEEcompsocthanksitem Jindong Wang is with William \& Mary. Email: jwang80@wm.edu.
\IEEEcompsocthanksitem Correspondence to: Jindong Wang (jwang80@wm.edu)

}

\thanks{Manuscript received March 5, 2024; revised Jan 6, 2025.}}

\markboth{Journal of \LaTeX\ Class Files,~Vol.~14, No.~8, August~2021}%
{Shell \MakeLowercase{\textit{et al.}}: A Sample Article Using IEEEtran.cls for IEEE Journals}

\IEEEpubid{0000--0000/00\$00.00~\copyright~2025 IEEE}

\IEEEtitleabstractindextext{%
\begin{abstract}
Foundation models are usually pre-trained on large-scale datasets and then adapted to different downstream tasks through tuning. 
This pre-training and then fine-tuning paradigm has become a standard practice in deep learning.
However, the large-scale pre-training datasets, often inaccessible or too expensive to handle, can contain label noise that may adversely affect the generalization of the model and pose unexpected risks. 
This paper stands out as the first work to comprehensively understand and analyze the nature of noise in pre-training datasets and then effectively mitigate its impacts on downstream tasks. 
Specifically, through extensive experiments of fully-supervised and image-text contrastive pre-training on synthetic noisy ImageNet-1K, YFCC15M, and CC12M datasets, we demonstrate that, while slight noise in pre-training can benefit in-domain (ID) performance, where the training and testing data share a similar distribution, it always deteriorates out-of-domain (OOD) performance, where training and testing distributions are significantly different.
These observations are agnostic to scales of pre-training datasets, pre-training noise types, model architectures, pre-training objectives, downstream tuning methods, and downstream applications. 
We empirically ascertain that the reason behind this is that the pre-training noise shapes the feature space differently.
We then propose a tuning method (NMTune) to affine the feature space to mitigate the malignant effect of noise and improve generalization, which is applicable in both parameter-efficient and black-box tuning manners, considering one may not be able to access or fully fine-tune the pre-trained models.
We additionally conduct extensive experiments on popular vision and language models, including APIs, which are supervised and self-supervised pre-trained on realistic noisy data for evaluation. 
Our analysis and results demonstrate the importance of this novel and fundamental research direction, which we term as \textit{Noisy Model \revision{Transfer} Learning}.
\end{abstract}

\begin{IEEEkeywords}
Noisy Model \revision{Transfer} Learning, Pre-training Noise, Parameter-efficient Tuning, Black-box Tuning
\end{IEEEkeywords}}

\maketitle

\IEEEdisplaynontitleabstractindextext

%
\IEEEpeerreviewmaketitle

\ifCLASSOPTIONcompsoc
\IEEEraisesectionheading{\section{Introduction}\label{sec:introduction}}
\else
\section{Introduction}
\label{sec:introduction}
\fi

\IEEEPARstart{F}{oundation} models \cite{bommasani2021opportunities} have become the standard practice of many applications in the rapidly evolving landscape of machine learning.
They have facilitated many machine learning tasks: instead of training a model from scratch for each individual task, which can be time-consuming, resource-intensive, and less adaptable, one can simply apply these foundation models or (partially) fine-tune \cite{kornb2019transfer,hu2021lora,he2021towards} them on various downstream tasks to achieve promising performance \cite{he2019moco,radford2021learning,he2022mae,brown2020language}.
The adaptation of these models in general follows the transfer learning paradigm known as pre-training and then fine-tuning (PT-FT) \cite{kornb2019transfer}, where they are first pre-trained using proxy training objectives on large-scale datasets that are crawled from web \cite{schuhmann2022laionb,together2023redpajama}, and then are adapted to downstream tasks.

For instance, ResNets \cite{he2015resnet}, Vision Transformers (ViT) \cite{dosovitskiy2020image,liu2021swin}, and ConvNext \cite{liu2022convnext} have been pre-trained in fully-supervised fashion on ImageNet-1K, ImageNet-21K, and larger but potentially more noisy datasets \cite{kolesnikov2020big,xie2020ns,ridnik2021imagenet21k,kakaobrain2022coyo700m}.
These models have then been applied in various downstream tasks of computer vision, such as detection and segmentation \cite{ren2015faster,lin2017focal,he2017mask,carion2020end,cheng2021per}. 
Contrastive language-image pre-trained (CLIP) models \cite{radford2021learning}, adopting a contrastive objective \cite{he2019moco} between image and text, present unprecedented zero-shot performance on downstream tasks. 
Language models, especially recent large language models such as GPT-4 \cite{openai2023gpt4} and Gemini \cite{gemini}, also follow this prevalent PT-FT paradigm \cite{devlin2018bert,liu2019roberta,radford2018improving,radford2019language,brown2020language,touvron2023llama}.

Over the years, there have been tremendous efforts in improving the performance of transferring the pre-trained models in various practical downstream scenarios, such as out-of-distribution generalization \cite{chen2021mandoline,kumar2022fine}, semi-supervised learning \cite{sohn2020fixmatch,usb2022}, imbalanced learning \cite{zhang2023deep, wang2023exploring}, noisy label learning \cite{song2022noisy,li2022selective}, to name a few. 
While it is essential for foundation models to deal with different tasks with learning and tuning techniques at downstream, it has also become a common belief recently that the distribution of pre-training data may play a more critical and straightforward role in the models' performance on downstream tasks \cite{Gadre2023DataCompIS,entezari2023role, zhang2023trade}.
The \emph{quality} of the pre-training data is also found more important for robust generalization compared to \emph{quantity} \cite{nguyen2022quality,lee2022dedup,gunasekar2023textbooks}.
However, less quantity and diversity of the pre-training data may not guarantee the reasonable performance of the foundation models \cite{longpre2023pretrainer}.
The bias in pre-training data created during the collection (and annotation) process, e.g., corrupted, poisoned, and false information \cite{blodgett2017racial,chang2020adversarial,birhane2023into,elazar2023s}, can also impose malicious and unexpected influence to downstream tasks \cite{bommasani2021opportunities,carlini2022quantifying,thiel2023identifying}.

This paper focuses on the label noise in large-scale pre-training datasets, which inevitably exists owing to the data collection process by human annotators and web crawlers and may adversely affect the transferability and generalization on downstream tasks of models pre-trained on such data.
For example, training CLIP \cite{radford2021learning} on LAION-2B \cite{schuhmann2022laionb}, which is a billion-scale uncurated image-text pair dataset, can just match the performance of training it on WIT-400M \cite{radford2021learning}, which is heavily and carefully cleaned and processed by OpenAI. 
Such noise can be extremely difficult to avoid or eliminate without sacrificing the quantity and diversity of data in the large-scale pre-training \cite{ridnik2021imagenet21k,vasudevan2022does,schuhmann2022laionb}.
While the exact relationship and modeling between the pre-training noise and downstream performance remain largely unknown, there are already numerous models pre-trained on large-scale noisy data and have been transferred on downstream tasks, such as Noisy Student \cite{xie2020ns}, BiT \cite{kolesnikov2020big}, and Open CLIP \cite{cherti2023reproducible}.
Not to mention the enormous but noisy raw text \cite{yang2019xlnet,lee2022dedup} that has been utilized to pre-train language models such as BERT \cite{devlin2018bert} and GPT \cite{radford2019language,brown2020language}. 
As the pre-trained models and datasets grow significantly, it has also become increasingly critical and challenging to understand \emph{how the noise in pre-training data affects the performance of pre-trained models on downstream tasks.} 
We present the first study on this important yet unexplored problem, demystifying the label noise in pre-training data, understanding its effects on downstream tasks, and then mitigating such (malignant) effects. 
Our study aims to answer the following key questions: 1) \emph{Influence:} How does the noise in pre-training data influence the downstream performance? 2) \emph{Analysis:} Why does such influence happen? and 3) \emph{Mitigation:} Based on the analysis, how to mitigate such influence on downstream tasks without re-training models from scratch?

\begin{figure*}[t!]
    \centering
    \includegraphics[width=0.9\linewidth]{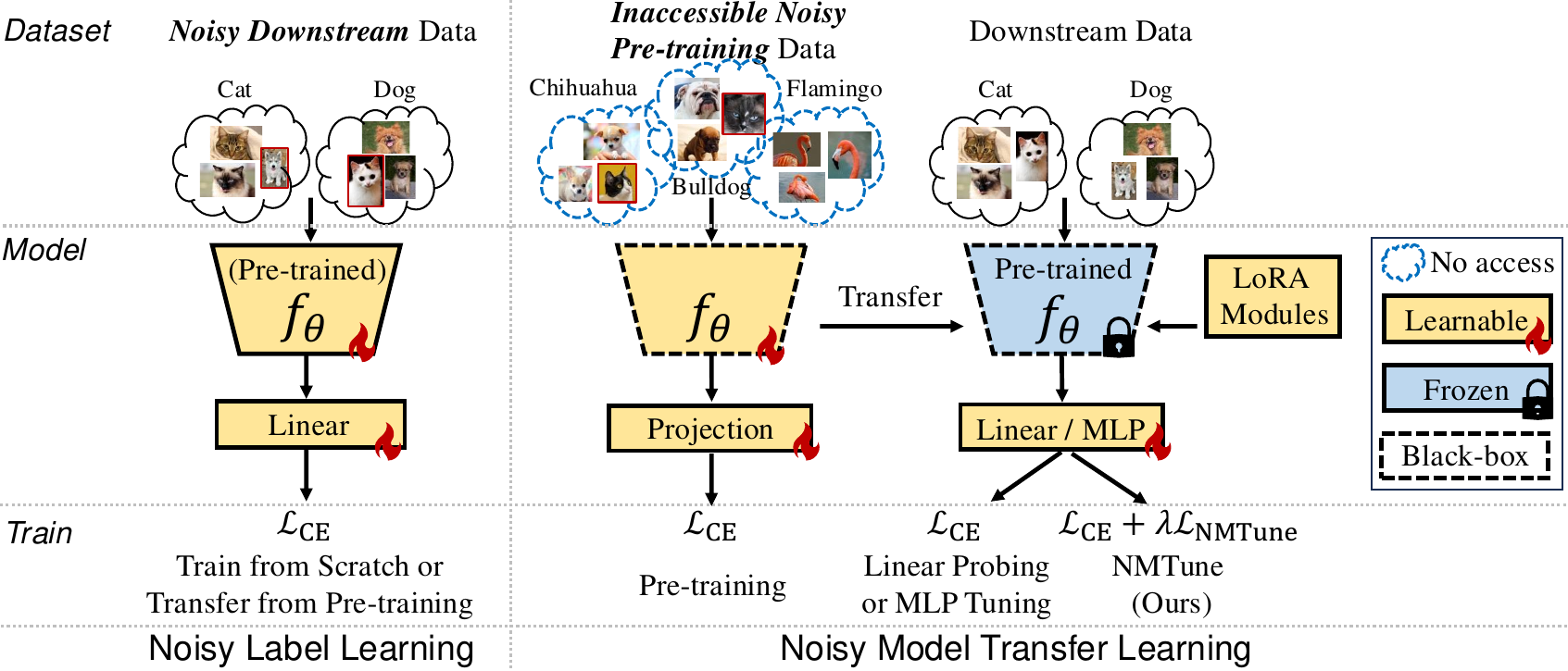}
    \vspace{-.05in}
    \caption{Illustration of noisy label learning (left) and the proposed \emph{noisy model \revision{transfer} learning} (right). Noisy label learning mainly focuses on robustly training a model from scratch or fine-tuning a model from pre-training on a noisy downstream dataset. Noisy model \revision{transfer} learning focuses on robustly adapting the (partially) black-box noisy pre-trained models to various downstream tasks, where we do not make additional assumption.}
    \label{fig:method}
\vspace{-0.1in}
\end{figure*}

Notably, there is a widely studied topic termed ``noisy label learning'' that studies training robust models \emph{given} noisy (downstream) training data \cite{Ghosh2017RobustLF, Li2020DivideMixLW, northcutt2021confidentlearning}.
Our problem is inherently different from noisy label learning, since we assume that the label noise exists in the usually inaccessible pre-training data but aims at understanding and mitigation of them on downstream tasks, as illustrated in \cref{fig:method}.
We identify this problem as a highly practical setting: due to the increasing size of pre-trained models and datasets, it becomes notoriously difficult to alter the pre-training process or fine-tune the entire models (black-box or cannot be updated due to the large parameter size and the constrained computation).
For instance, the open-source Llama \cite{touvron2023llama,touvron2023llamav2} model requires multiple NVIDIA V100 or A100 GPUs to fine-tune and even infer, which is not affordable for most ordinary researchers.
Moreover, proprietary models like ChatGPT only provide the API, which cannot be locally fine-tuned and diagnosed to fix the adverse effect of pre-training noise. 
Sometimes, the pre-training data is proprietary or black-box, which further introduces difficulty in mitigating noise and thus requires block-box techniques to overcome the influence of noise in pre-training on downstream tasks.


\begin{figure}[t]
    \centering
    \includegraphics[width=0.98\linewidth]{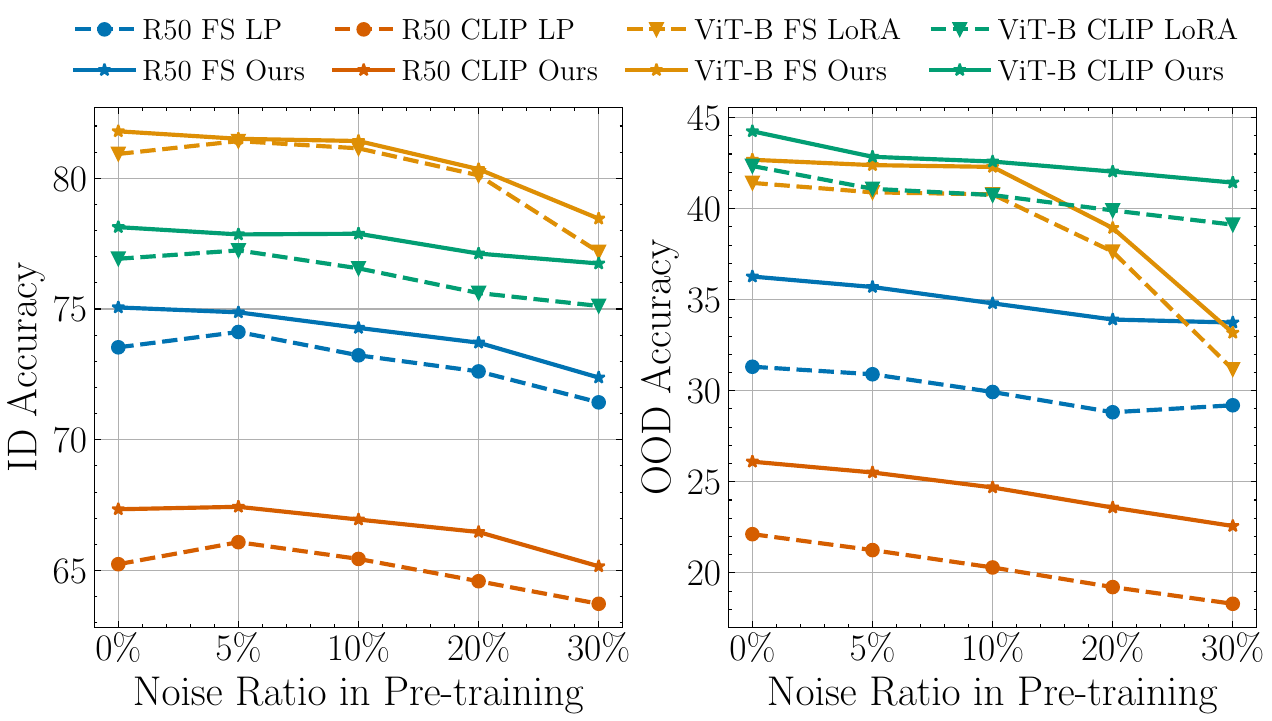}
    \caption{In-domain (ID) and out-of-domain (OOD) downstream performance when fully-supervised (FS) pre-training the model on synthetic noisy ImageNet-1K (IN-1K) and image-text contrastive pre-training YFCC15M (and CC12M) of ResNet-50 and ViT-B-16 on various noise ratios. 
    We compare linear probing (LP) and parameter-efficient tuning of LoRA, with the proposed method. 
    On ID, $5\%$ noise in pre-training benefits the LP performance. Our method not only boosts the general performance but also rectifies the model pre-trained on clean data to be comparable to $5\%$ noise. 
    On OOD, noise in pre-training is detrimental to robustness performance when conducting LP and LoRA.  
    Our method improves the transferability on OOD tasks significantly.}
    \label{fig:teaser}
\vspace{-0.1in}
\end{figure}

This paper extends our previous paper accepted at ICLR 2024 \cite{chen2023understanding}, with more analysis of the influence of the pre-training noise using larger models, pre-training of larger scale, different noise types, different tuning methods, and on downstream tasks other than classification.
More specifically, we present in-depth experiments to answer the above questions, based on the popular \emph{fully-supervised} (FS) and \emph{image-language contrastive} (i.e., CLIP) pre-training. 
\begin{itemize}
    \item \textbf{Influence: The label noise in pre-training data has both benevolent and malignant influence on downstream tasks.} In \cref{sec:understand}, we conduct realistic and practical experiments with ResNet-50 \cite{he2015resnet} and ViT-B-16 \cite{dosovitskiy2020image} models fully-supervised and contrastive pre-trainied on synthetic noisy ImageNet-1K, YFCC15M \cite{2016yfcc100m}, and CC12M \cite{changpinyo2021cc12m} with various noisy ratios $\{0\%, 5\%, 10\%, 20\%, 30\%\}$ and then study the generalization performance on the downstream in-domain (ID) and out-of-domain (OOD) classification tasks, object detection, and instance segmentation tasks. We observe that, on ID tasks, slight noise (up to $5\%$ or $10\%$) can benefit generalization performance. In contrast, even $5\%$ noise can drastically deteriorate robustness and transferability on OOD tasks, as shown in \cref{fig:teaser} and \cref{fig:r50_in_ood_eval}.
    One can consistently find these observations on downstream tasks of different applications, tuning methods \cite{he2021towards,hu2021lora}, and w.r.t. pre-training noise types.
    \item \textbf{Analysis: Label noise in pre-training shapes the feature space significantly of the pre-trained model.} 
    In \cref{sec:analysis-empirical}, 
    we conduct analysis from the singular value spectrum on the feature space of the pre-trained models. Noise in pre-training results in the decreasing largest singular value and flatter singular value distribution with a higher dimension span in the feature space.
    An initial increase in the spanning dimension of the feature space is beneficial for discriminability in ID tasks. Still, it then becomes detrimental with the further increase, indicating that more feature space capacities are learned to fit the noise structure. The decrease in dominant singular values leads to less transferability for OOD tasks \cite{chen19itrans}, as shown in \cref{fig:r50_id_ood_svd}.
    \item \textbf{Mitigation: We design a simple tuning algorithm to reshape the pre-trained feature space, reducing the influence of noisy pre-training data and boosting the performance of downstream tasks.}
    Our method is light-weight and compatible in both black-box and parameter-efficient tuning paradigms, as shown in \cref{fig:method}.
    In \cref{sec:mitigate}, based on the analysis, we propose three regularization objectives on the singular value spectrum that help affine the feature space. We demonstrate the effectiveness of the proposed method on noisy ResNet-50 and ViT-B-16 models with extensive analysis, as shown in \cref{fig:teaser} and \cref{fig:r50_results}.
    In \cref{sec:exp}, we further validate our method on popular noisy pre-trained models with superior generalization performance for both vision and language tasks.
\end{itemize}
Beyond our analysis, we view this research as a novel and complementary topic to the classic noisy label learning setting, termed as \emph{Noisy Model \revision{Transfer} Learning} (NML).
We think the value of this direction is even more significant in the era of large foundation models \cite{bommasani2021opportunities}, where the downstream users only have access to model weights or APIs. 
It would be of particular interest to explore how to eliminate the malignant influence of noise in pre-training on downstream tasks when adapting these models without full fine-tuning, since it may exist in broader applications such as the detection and segmentation in medical and autonomous driving.
We hope that future research on this topic can facilitate a better understanding and application of large foundation models.

\section{Related Work}
\label{sec:related}

\subsection{Noisy Label Learning}
Noisy label learning has emerged as a critical area of focus within machine learning research, given its significance in improving model robustness against inaccurately labeled data. 
The field has seen extensive studies aiming at developing methods and techniques to either train models robust to noise from the ground up or to adapt models, initially trained on clean datasets, to perform effectively on noisy (downstream) datasets. 
These efforts encompass a variety of strategies, including the design of robust loss functions \cite{Ghosh2017RobustLF, Zhang2018GeneralizedCE, Wang2019SymmetricCE, Ma2020NormalizedLF}, the estimation of label noise levels \cite{Xiao2015LearningFM,Goldberger2016TrainingDN,Liu2020EarlyLearningRP,northcutt2021confidentlearning,ICML:Li+etal:2021}, and the correction of noisy labels \cite{Han2018CoteachingRT, Li2020DivideMixLW, zhang2021learning,sopliu22w, kim2021fine,chen2023imprecise}.

A particularly relevant topic of research within this domain has been the theoretical exploration of the conditions under which a loss function can exhibit tolerance to noisy labels, as extensively analyzed by Ghosh et al. \cite{Ghosh2017RobustLF}. 
While earlier studies predominantly focused on synthetic label noise, more recent works have sought to understand and model real-world, instance-dependent label noise. 
For instance, CIFAR-N \cite{wei2021learning} was developed as a benchmark to facilitate the study of label noise as it occurs in real-world scenarios, marking a significant step towards bridging the gap between theoretical models and practical applications.
Furthermore, the work by Cheng et al. \cite{cheng2021mitigating} introduced innovative techniques to mitigate the memorization of noisy labels by analyzing and manipulating the regularization between data representations. 
This approach underscores the importance of representation learning in combating the adverse effects of label noise. 
Similarly, Wen et al. \cite{wen2022benign} provided a rigorous theoretical analysis that illustrates the challenges posed by benign overfitting in the presence of label noise, shedding light on the limitations of traditional overfitting paradigms under noisy conditions.
Xue et al. \cite{xue2022investigating} expanded the scope of investigation to include the robustness of contrastive pre-training methods when faced with noisy labels, particularly in the context of downstream tasks. Their work not only explores the effectiveness of pre-training strategies in noise mitigation but also delves into the transferability of learned representations, using the singular value spectrum of the pre-trained feature space as a metric for analysis.

In contrast to the aforementioned studies, our work introduces a novel perspective by focusing on the impact of label noise during the \emph{pre-training} phase and its subsequent effects on a variety of downstream tasks. This approach not only differentiates our research from the traditional noisy label learning paradigm but also complements it by providing insights into how pre-training under noisy conditions can influence model performance across different tasks.



\subsection{Pre-training and Fine-Tuning}

Pre-training and fine-tuning represent the cornerstone of the transfer learning paradigm, enabling the adaptation of models pre-trained on a large dataset to perform effectively on new, albeit similar, datasets. 
This approach has proven particularly beneficial in leveraging learned representations from a broad data spectrum to enhance performance on more specific tasks. 
Various techniques have been proposed to address challenges such as distribution shifts \cite{cheng2021mitigating}, the integration of unlabeled data \cite{sohn2020fixmatch,zhang2021flexmatch,wang2022freematch}, handling imbalanced datasets \cite{kang2019decoupling,wang2023margin}, and mitigating the impact of noisy data \cite{wei2021smooth,xue2022investigating}. 
These methods underscore the adaptability of transfer learning strategies in diverse data conditions.

In addition to these techniques, significant research efforts have been directed towards optimizing the pre-training process itself to further improve transfer performance. This includes exploring the trade-off between diversity and specificity in pre-training data \cite{kaplan2020scaling,zhang2023trade}, selective data sampling for pre-training \cite{entezari2023role}, and the balance between the data quality and quantity \cite{magar2022data,nguyen2022quality,lee2022dedup,carlini2022quantifying,Gadre2023DataCompIS}. Such studies aim to refine the foundational datasets for pre-training, enhancing the model's ability to generalize and subsequently adapt to new tasks. Furthermore, specialized fine-tuning methodologies have been developed \cite{tsai2020transfer,kumar2022fine,wortsman2022robust,goyal2023finetune,xu2023towards} to tailor pre-trained models more effectively to specific downstream tasks, illustrating the depth of research focused on optimizing each phase of the transfer learning process.

Parameter-efficient transfer learning methods \cite{he2021towards,oh2023blackvip}, such as the use of adapters \cite{houlsby2019parameter}, Low Rank Approximation (LoRA) \cite{hu2021lora}, and prompt tuning \cite{liu2022p,liu2021p}, represent a shift towards more light-weight and more adaptable transfer learning frameworks, especially for large foundation models. 
These approaches allow for significant customization and efficiency improvements by modifying only a small fraction of the model's pre-trained parameters, thereby preserving the underlying pre-trained model's integrity while achieving customization for specific downstream tasks. 

Notably, while there is a rich body of work assuming the availability and modifiability of pre-trained models, our research navigates the less explored territory of applying transfer learning principles to (partially) black-box models. This distinction is critical as it addresses a practical constraint in many real-world applications where the internal structures may not be accessible for direct modification.



\subsection{Pre-training Data Biases}

The rapid evolution of foundation models, alongside the exponential growth in the size of datasets used for pre-training, has introduced complex challenges related to data quality and bias. As these datasets grow larger, the inclusion of low-quality samples becomes more probable \cite{elazar2023s}, leading to issues such as skewed data distributions \cite{kandpal2023large} and the incorporation of private or unethical content \cite{genderbiasllm,omiye2023large}. Such biases in pre-training data, once learned by foundation models, can inadvertently transfer and inherit to downstream tasks, potentially resulting in a range of negative outcomes. These can include serious security and privacy vulnerabilities \cite{wei2023jailbroken,zou2023universal}, degraded model generalization capabilities \cite{recht2019imagenet,zhang2023trade}, and the emergence of undesirable behaviors \cite{Nakkiran2019DeepDD}.

The ramifications of pre-training data biases are profound, affecting not just model performance, but also the ethical and social implications of deploying these models in real-world applications. Concerns over these biases have led to a growing body of empirical research aimed at evaluating the impact of various factors, such as toxicity levels, data quality, and chronological aspects of the pre-training data on the downstream effectiveness of large language models \cite{longpre2023pretrainer}. These studies are crucial for developing a nuanced understanding of how the nature pre-training data influence model behavior in diverse application contexts.
Our study embarks on a pioneering investigation into the specific impacts of noise within pre-training data on downstream task performance of different applications.

\section{Noisy Model \revision{Transfer} Learning}
\label{sec-nml}

In this section, we formally define the problem setting and learning objective of noisy model \revision{transfer} learning (NML).
We assume that a foundation model $f_{\phi}$ with learnable parameters $\phi$ is pre-trained using a proxy objective/algorithm $\mathcal{A}_{\mathrm{up}}$ on a large-scale but noisy dataset $\hat{\mathcal{D}}_{up}$. 
After pre-training, we assume that the foundation model cannot be fully accessible and we can only perform black-box or parameter-efficient tuning $\mathcal{A}_{\mathrm{down}}$ with learnable modules $g_\theta$ on downstream tasks with dataset $\mathcal{D}_{\mathrm{down}}$. 
Note that $g_\theta$ can be of many forms, such as a linear classifier in black-box tuning or the light-weight modules on LoRA in parameter-efficient tuning. 
\begin{definition}
    NML aims to understand and model the relationship between the noise in the pre-training data $\hat{\mathcal{D}}_{up}$, which is learned by the foundation model $f_{\phi}$, and the downstream performance by tuning this foundation:
    \begin{equation}
        g_{\theta}\left( \mathcal{D}_{\mathrm{down}}, f_{\phi}\left(\hat{\mathcal{D}}_{\mathrm{up}}, \mathcal{M}, \mathcal{A}_{\mathrm{up}} \right), \mathcal{A}_{\mathrm{down}} \right),
    \end{equation}
    where $\mathcal{M}$ denotes the foundation model architecture. 
    NML also includes the mitigation of the detrimental effects of pre-training noise on downstream performance with $\mathcal{A}_{\mathrm{down}}$.
\end{definition}
\noindent 
\revision{In this work, 'understand' refers to empirically and theoretically analyzing how pre-training noise influences the feature space, learned representations, and generalization performance on pre-training and downstream tasks. 
This involves techniques such as feature space analysis (e.g., singular value spectrum analysis in Section 5) and performance evaluations under varying noise conditions. 
'Model' refers to developing formal approaches or algorithms (e.g., the proposed NMTune method in Section 6) that address and mitigate the impact of pre-training noise during downstream adaptation.}
NML differs from noisy label learning (NLL), not only in that NML focuses on the pre-training noisy datasets $\hat{\mathcal{D}}_{up}$ whereas NML focuses on downstream noisy datasets $\hat{\mathcal{D}}_{\mathrm{down}}$, but also we assume $f_{\phi}$ is (partially) black-box for us at the downstream tasks. 
Thus, it is challenging to understand and mitigate the pre-training noise on downstream tasks. 

\section{Understanding the Pre-training Noise}
\label{sec:understand}

In this section, we empirically and systemically investigate the effect of noisy labels on the learned representations in the supervised pre-training. 
We build our evaluation on realistic and practical motivating experiments of training ResNet-50 \cite{he2015resnet} and ViT-B-16 \cite{dosovitskiy2020image} on the synthetic noisy ImageNet-1K \cite{ILSVRC15}, YFCC15M (a subset of YFCC100M \cite{2016yfcc100m}), and CC12M \cite{changpinyo2021cc12m}.
To fully understand the effect of pre-training noise on downstream tasks, we also study different tuning methods and various downstream applications.

\subsection{Experiments Design}
\label{sec:exp-design}

Here, we first introduce the general pipeline for our motivating experiments. 
We pre-train the models on noisy pre-training datasets by manually introducing synthetic supervision noise, and then fine-tune and evaluate the pre-trained noisy models on different downstream applications.

\textbf{Noisy pre-training datasets}. 
We assume the pre-training dataset consists of inputs $\mathbf{x} \sim \mathcal{X} $ and supervision $y \sim \mathcal{Y}$.
We define a clean dataset $\mathcal{D} = \{(\mathbf{x}_i, y_i)\}_{i \in [N]}$ of size $N$ with accurate supervision, where $[N]:=\{1,\ldots,N\}$. 
We assume that $y$ can exist in different formats in pre-training, e.g., an actual label for the input as in fully-supervised learning \cite{ILSVRC15, he2015resnet, ridnik2021imagenet21k} or a text description for an input image as in image-text contrastive learning of CLIP \cite{2016yfcc100m, radford2021learning, jia2021scaling, changpinyo2021cc12m, desai2021redcaps, schuhmann2021laion400m, schuhmann2022laionb}.
Due to the scale of data collection and the cost of data annotation, the pre-training dataset can usually contain noisy supervision $\hat{y}$ that does not accurately match the corresponding $\mathbf{x}$ \cite{recht2019imagenet, beyer2020imagenet, northcutt2021confidentlearning, yun2021relabel, vasudevan2022does, schuhmann2022laionb}. 
We define such a noisy pre-training dataset as $\hat{\mathcal{D}} = \{(\mathbf{x}_i, \hat{y}_i)\}_{i \in [N]}$ and correspondingly the noise ratio $\gamma$ as the percentage of noisy supervision in $\mathcal{\hat{D}}$. 
In this section, we only consider the random noise of all concepts/classes, and we provide discussion of asymmetric noise in \cref{sec:asym-noise}.

\textbf{Pre-trained models}. 
The pre-trained models serve as a foundation for various downstream tasks and usually can be abstracted as the stack of a feature extractor and a projection head. 
We define the feature extractor with learned parameters $\phi$ as a mapping function $f_{\phi}$ from the input space to the feature space of dimension $D$: $f_{\phi}: \mathcal{X} \rightarrow \mathcal{F}$.  
The projection head $g_{\theta}: \mathcal{F} \rightarrow \mathcal{Y}$ is jointly pre-trained with the feature extractor using the proxy pre-training objective, and thus is usually not used when adapting $f_{\phi}$ on downstream tasks. 
We consider two types of supervised pre-training paradigms on images for our motivating experiments: fully-supervised pre-training where $y$ is the actual class label and the projection head is a linear classifier \cite{he2015resnet}, and contrastive pre-training with text supervision (CLIP) where $y$ is the text description given an input image and the projection is a multi-layer perceptron (MLP) non-linear function mapping the image and text to a common feature space \cite{radford2021learning,cherti2023reproducible}.

\textbf{Downstream in-domain (ID) and out-of-domain (OOD) evaluation of classification tasks}.  
To investigate the effect of noisy supervision comprehensively, we leverage both in-domain (ID) and out-of-domain (OOD) evaluation to assess the generalization capability of the pre-trained feature extractor \cite{josip2021cnnrobusttransfer}  $f_{\phi}^{\gamma}$ that are obtained from the pre-training data of different noise ratios.
To evaluate the pre-trained models on a downstream dataset $\mathcal{D}' = \{(x_i, y_i)\}_{i \in [M]}$ of downstream classification task, where we can always treat $y \in [C]$ as an actual class label and do not make assumption whether it is clean or noisy, and measure the quality of the learned representation, we conduct linear probing (LP), where only a $C$-way linear classification head is re-trained on the downstream dataset and the feature extractor is frozen.
Linear probing is a common evaluation protocol accessing feature quality \cite{he2019moco,liu2021selfsupervised} that can be viewed as a simple black-box tuning method for pre-trained models that are typically large and difficult or unable to fully fine-tune.
Since most of the large foundation models nowadays are usually adapted with parameter-efficient tuning \cite{he2021towards}, we also mimic this situation and evaluate the pre-trained noisy models by partially tuning only a small proportion of their learnable parameters via LoRA \cite{hu2021lora}. 
Compared to LP, LoRA is a white-box tuning method that inserts light-weight learnable modules at intermediate layers of the pre-trained models.
In addition, to fully understand the effect of noise on pre-trained models, we also conduct full fine-tuning of pre-trained models.
For ID evaluation, we assume the same marginal distribution on $\mathcal{X}$ for both training and testing. 
In contrast, for OOD evaluation, we tune the models on a source distribution and evaluate them on different (multiple) target distributions \cite{kumar2022fine}.

\textbf{Downstream evaluation on detection and segmentation tasks}. In addition to classification, we further evaluate pre-trained models on downstream object detection and instance segmentation tasks \cite{lin2014microsoft,ren2015faster,lin2017focal,he2017mask}. 
In object detection, the downstream task is to regress the bounding boxes of objects in the input image, and the supervision $y$ becomes the location of these bounding boxes. 
Instance segmentation is a pixel-level classification task and $y$ is thus the pixel mask of each instance in the input image. 

\subsection{Pre-training Results}

For pre-training, we use both ResNet-50 \cite{he2015resnet} and ViT-B-16 \cite{dosovitskiy2020image} for convolutional neural network \cite{lecun1998gradient} and transformer \cite{vaswani2017attention} architectures, respectively. 
Two pre-training proxy tasks that require supervision are adopted: fully-supervised and image-text contrastive pre-training of different scales.

\textbf{Fully-supervised (FS) pre-training setup}. 
We use ImageNet-1K (IN-1K) \cite{ILSVRC15} for fully-supervised pre-training. 
To introduce noisy supervision in the datasets, we uniformly flip the ground truth class label into the \revision{another randomly sampled classes} in IN-1K using cleanlab \cite{northcutt2021confidentlearning} by setting the noise ratio $\gamma$ to $\{0\%, 5\%, 10\%, 20\%, 30\%\}$, where $0\%$ denotes no noise (i.e., the original dataset). 
For ResNet-50, we follow the IN-1K pre-training recipe in \cite{wightman2021resnet}, where the models are trained for 600 epochs using binary cross-entropy (BCE) loss on 224$\times$224 images. 
The global batch size is set to 2048 and the LAMB optimizer \cite{you2019large} is used. 
We set the learning rate to 5$e$-3, which is annealed according to the cosine scheduler with 5 warm-up epochs, and the weight decay is $0.01$.
We use label smoothing of $0.1$ \cite{He2019CVPR}, stochastic depth of $0.05$ \cite{huang2016deep}, repeated augmentation \cite{berman2019multigrain} consisting of RandAugment \cite{cubuk2020randaugment}, Mixup \cite{zhang2017mixup}, and CutMix \cite{yun2019cutmix}. 
For ViT-B-16, we follow the training recipe mainly in \cite{touvron2022deit}.
We use a batch size of 4096, LAMB optimizer with a learning rate of 5$e$-3, and a weight decay of $0.02$.
All ViT-B-16 models are trained for 300 epochs with 5 warm-up epochs and BCE loss.
Similarly, we used a stochastic depth of $0.2$ and repeated augmentation. 
We do not use label smoothing, but we adopt LayerScale \cite{touvron2021going} and gradient clipping at $1.0$.

\textbf{Image-text contrastive (CLIP) pre-training setup}. 
We mainly follow \cite{cherti2023reproducible} for CLIP pre-training. 
Since vision transformers usually require more data to generalize well \cite{dosovitskiy2020image,cherti2023reproducible}, compared to convolutional neural networks, we train ResNet-50 on YFCC15M \cite{2016yfcc100m} and ViT-B-16 on both YFCC15M and CC12M \cite{changpinyo2021cc12m}, respectively.
For image-text pair datasets, we randomly swap the text description of two sampled pairs to introduce noise. 
Similarly, the noise ratio is set to $\{0\%, 5\%, 10\%, 20\%, 30\%\}$.
We set the learning rate for ResNet-50 and ViT-B-16 as 1$e$-3 and 5$e$-4, respectively, with a batch size of $4096$ and a weight decay of $0.1$. 
Both models are trained using image-text contrastive loss for 32 epochs.

\textbf{Results}. 
We present the IN-1K validation accuracy of the pre-trained models in \cref{tab:r50-imagenet-acc} to validate the pre-training recipes we used.
For IN-1K fully-supervised pre-trained models, we report model accuracy directly, whereas for CLIP pre-trained models, we report zero-shot accuracy \cite{radford2021learning,cherti2023reproducible}. 
Pre-trained models have performance comparable to publicly released models \cite{wightman2021resnet,touvron2022deit,cherti2023reproducible}. 
Interestingly, from the results, we found that ViT-B-16 is less robust to the pre-training \revision{label noise} compared to ResNet-50 in terms of pre-training evaluation, where, as the noise ratio in the pre-training dataset increases, the pre-training performance of ViT-B-16 degrades more than ResNet-50. 

\begin{table}[t!]
\centering
\caption{Validation accuracy of fully-supervised (FS) and image-text contrastive (CLIP) clean and noisy pre-trained ResNet-50 and ViT-B-16 models. }
\label{tab:r50-imagenet-acc}
\vspace{-0.1in}
\resizebox{0.85 \linewidth}{!}{%
\begin{tabular}{@{}c|cc|cc@{}}
\toprule
\multirow{2}{*}{\begin{tabular}[c]{@{}c@{}}Noise\\ Ratio \\ (\%)\end{tabular}} & \multicolumn{2}{c|}{ResNet-50 Acc. (\%)} & \multicolumn{2}{c}{ViT-B-16 Acc. (\%)}                                    \\ \cmidrule(l){2-5} 
                                                                               & IN-1K FS          & YFCC15M CLIP         & IN-1K FS & \begin{tabular}[c]{@{}c@{}}YFCC15M+\\ CC12M CLIP\end{tabular} \\ \midrule
0                                                                              & 79.96             & 32.64                & 78.70    & 45.43                                                         \\
5                                                                              & 79.18             & 30.86                & 78.13    & 44.10                                                         \\
10                                                                             & 78.61             & 29.54                & 77.19    & 43.22                                                         \\
20                                                                             & 76.27             & 27.72                & 74.67    & 40.48                                                         \\
30                                                                             & 73.11             & 26.53                & 68.12    & 38.57                                                         \\ \midrule
Public                                                                             & 80.04 \cite{wightman2021resnet}             & 30.18 \cite{cherti2023reproducible}                & 80.90 \cite{touvron2022deit}    & -                                                        \\ \bottomrule
\end{tabular}
}
\vspace{-0.2in}
\end{table}

\subsection{Results on Downstream Classification Tasks}

After obtaining the clean and noisy pre-trained models, we conduct different fine-tuning experiments on downstream classification tasks.
We perform both ID and OOD evaluation to comprehensively evaluate both the transferability and robustness of the pre-trained models.
We adopt three standard fine-tuning approaches: linear probing (LP), low-rank adaptation (LoRA~\cite{hu2021lora}) for parameter-efficient fine-tuning (of ViT-B-16), and full fine-tuning (FT).

For ID evaluation, we use $14$ downstream datasets including CIFAR-10/100 \cite{krizhevsky2009learning}, Flowers102 \cite{nilsback2008automated}, Food101 \cite{bossard14}, OxfordPet \cite{parkhi12a}, StanfordCars \cite{jonathan2013cars}, FGVCAircraft \cite{maji2013finegrained}, SVHN \cite{2011svhn}, DTD \cite{cimpoi14describing}, Caltech101 \cite{FeiFei2004LearningGV}, EuroSAT \cite{helber2018introducing,helber2019eurosat}, PatchCamelyon \cite{Veeling2018qh}, RESISC45 \cite{Cheng2017resic}, and Rendered SST2 \cite{socher2013recursive}, which cover various visual domains. 
For OOD evaluation, we use the ``real'', ``sketch'', ``inpainting'', and ``clippart'' of DomainNet \cite{peng2019moment}, where we train on either ``real'' or ``sketch'' and evaluate on the others.
For CLIP pre-trained models, we use 6 ImageNet variants for evaluation, including ImageNet-Adv (IN-A) \cite{dan2021nae}, ImageNet-R (IN-R) \cite{dan2021ood}, ImageNet-Sketch (IN-S) \cite{wang2019learning}, ImageNet-ViD (IN-V) \cite{vaishaal2021time}, ImageNet-V2 (IN-V2) \cite{recht2019imagenet}, and ObjectNet \cite{objectnet2019} for OOD evaluation while training on ImageNet-1K.

\subsubsection{Downstream results using linear probing}

\textbf{Setup.}
To perform LP, we freeze all the layers of the pre-trained models and fine-tune only the linear classifier that is randomly initialized on the top of the frozen backbone. 
We train the linear classifier for 30 epochs on each downstream dataset, using the AdamW \cite{kingma2014adam,loshchilov2017decoupled} optimizer with a cosine scheduler. 
We do not use weight decay for linear probing and set the learning rate at $0.01$ for all tasks. 
Each experiment is run with three random seeds.
We report the LP performance using $\{10\%, 25\%, 50\%, 75\%, 100\%\}$ percentage of downstream datasets to further evaluate the effect of pre-training noise w.r.t. number of data points.

\begin{figure*}[t!]
\centering
    \hfill
    \subfloat[ResNet-50 FS, ID]{\label{fig:r50_in1k_id_eval}\includegraphics[width=0.24\linewidth]{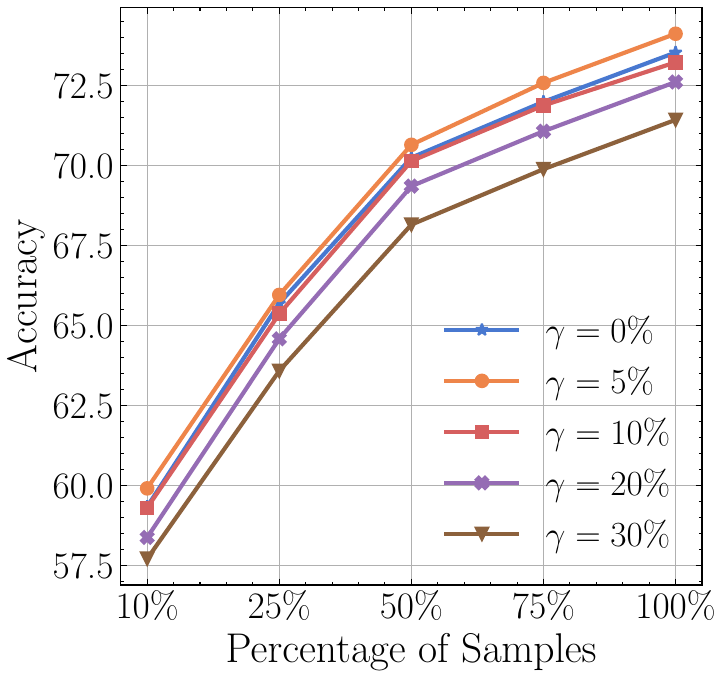}}
    \hfill
    \subfloat[ResNet-50 FS, OOD]{\label{fig:r50_in1k_odd_eval}\includegraphics[width=0.24\linewidth]{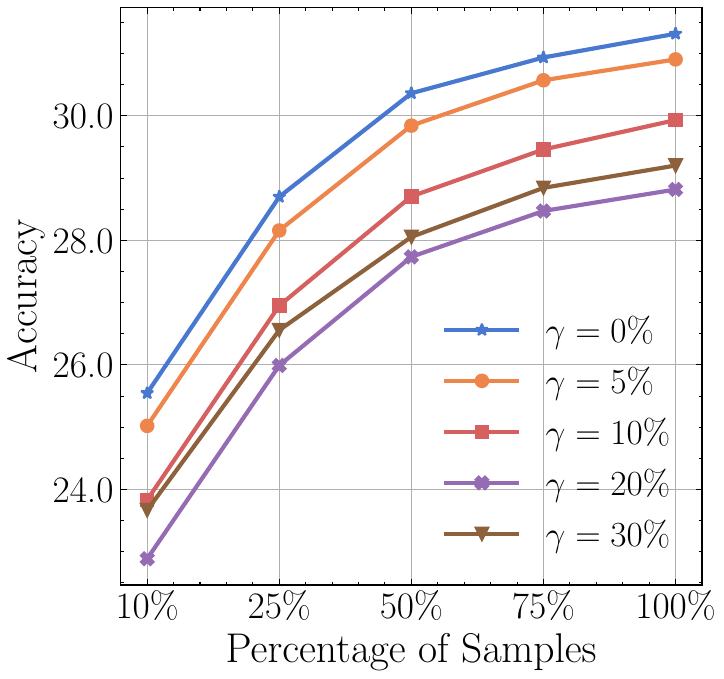}}
    \hfill
    \subfloat[ResNet-50 CLIP, ID]{\label{fig:r50_yfcc15m_id_eval}\includegraphics[width=0.24\linewidth]{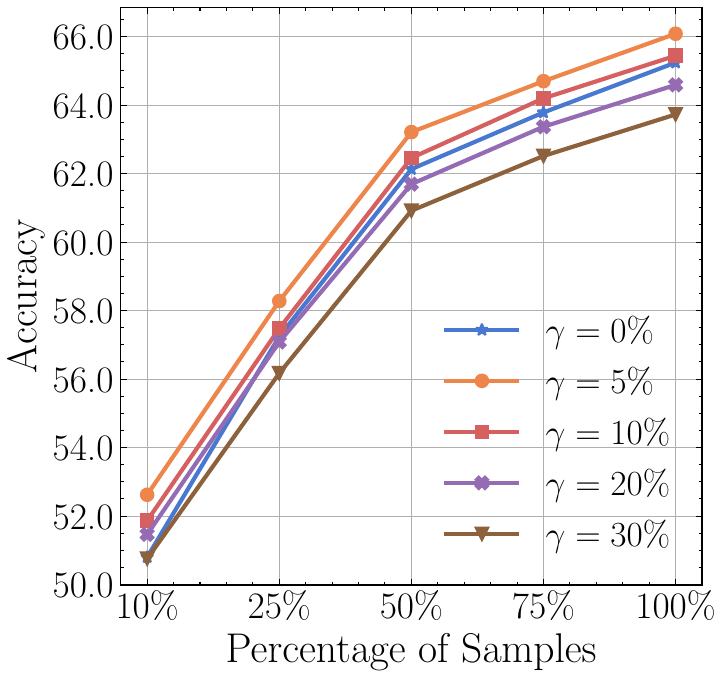}}
    \hfill
    \subfloat[ResNet-50 CLIP, OOD]{\label{fig:r50_yfcc15m_ood_eval}\includegraphics[width=0.24\linewidth]{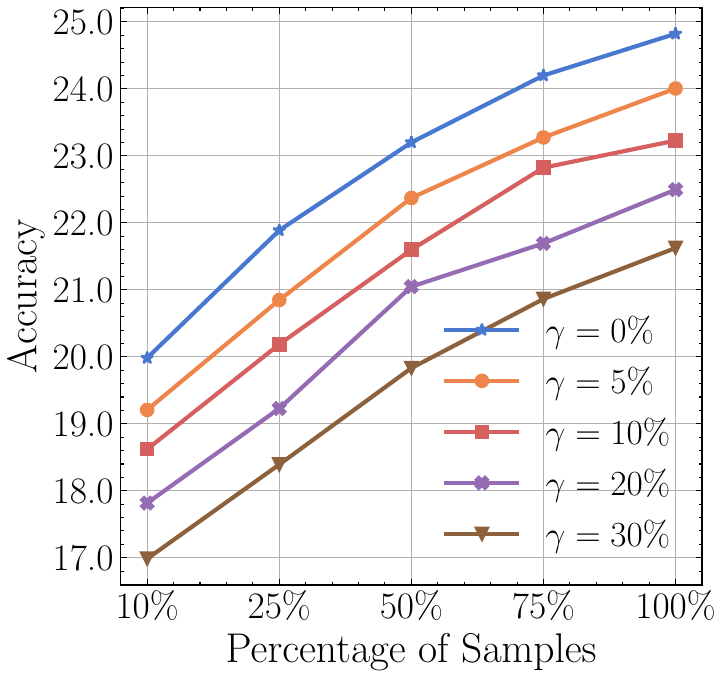}}
    \hfill

    \hfill
    \subfloat[ViT-B-16 FS, ID]{\label{fig:vitb16_in1k_id_eval}\includegraphics[width=0.24\linewidth]{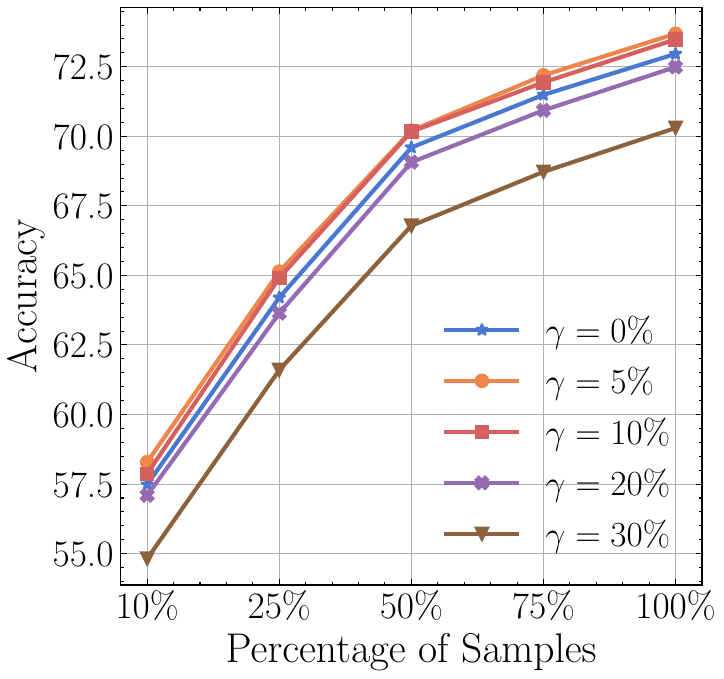}}
    \hfill
    \subfloat[ViT-B-16 FS, OOD]{\label{fig:vitb16_in1k_odd_eval}\includegraphics[width=0.24\linewidth]{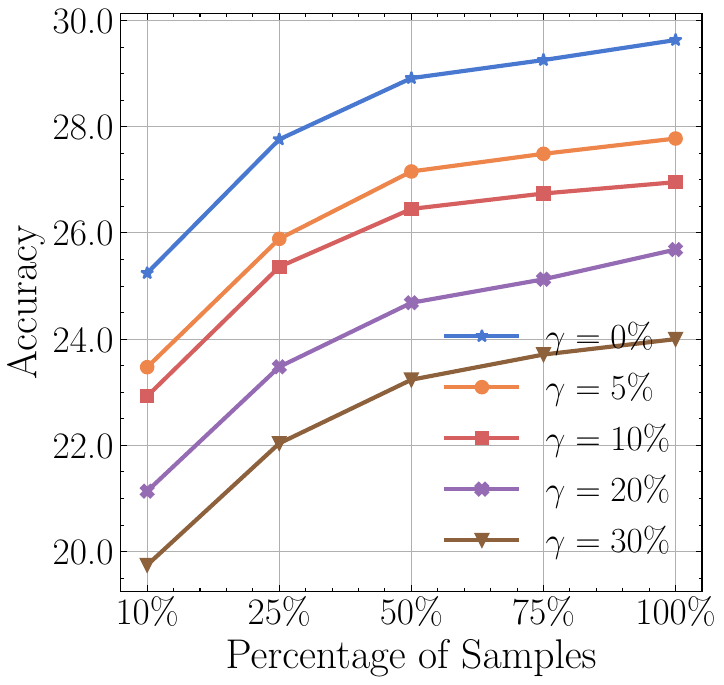}}
    \hfill
    \subfloat[ViT-B-16 CLIP, ID]{\label{fig:vitb16_yfcc15m_id_eval}\includegraphics[width=0.24\linewidth]{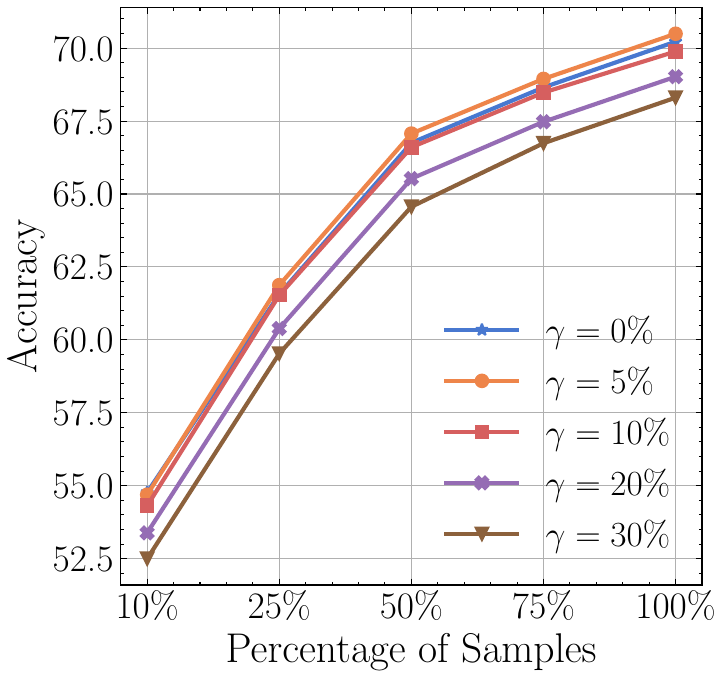}}
    \hfill
    \subfloat[ViT-B-16 CLIP, OOD]{\label{fig:vitb16_yfcc15m_ood_eval}\includegraphics[width=0.24\linewidth]{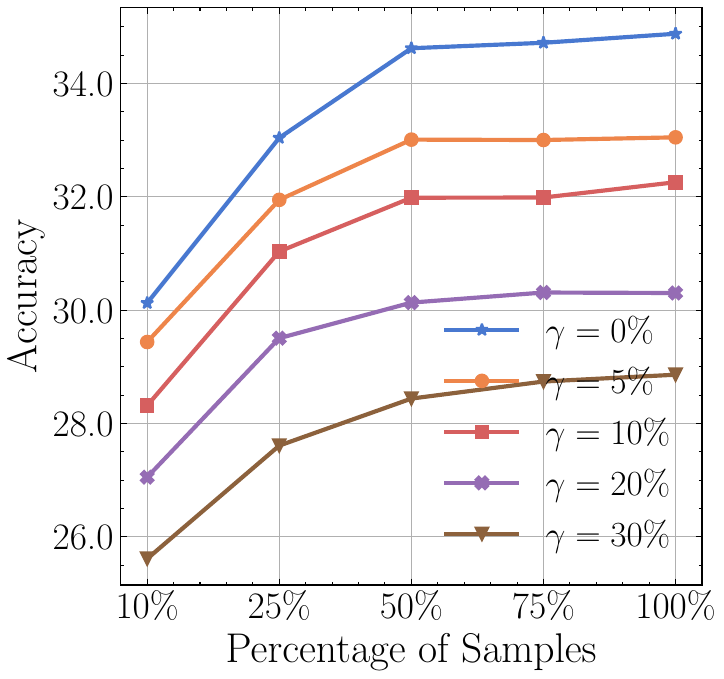}}
    \hfill

\caption{Average ID and OOD evaluation results of ResNet-50 (top row) and ViT-B-16 (bottom row), using ImageNet-1K (IN-1K) fully-supervised pre-training ((a), (b), (e), (f)) and YFCC15M (and CC12M) CLIP pre-training ((c), (d), (g), (h)) on downstream tasks with various percentages of data. 
For both ResNet-50 and ViT-B-16 pre-trained on datasets of different scales, on ID evaluation, the transferring performance first increases as noise increases (to $5\%$ or $10\%$) and then decreases with more noise. On OOD evaluation, the robustness performance constantly decreases. }
\label{fig:r50_in_ood_eval}
\vspace{-0.15in}
\end{figure*}

\textbf{Results}.
The average accuracy of both IN-1K FS and CLIP pre-trained ResNet-50 and ViT-B-16 is shown in \cref{fig:r50_in_ood_eval}. 
With extensive motivating experiments on different architectures and various pre-training data scales, we empirically find two important and surprisingly counter-intuitive observations from the LP results: 
\begin{mdframed}
\begin{itemize}[leftmargin=1em]
\setlength\itemsep{0em}
    \item Proper noisy supervision in pre-training (e.g., $5\%$ or $10\%$) can benefit the performance on ID downstream tasks, while more noise results in inferior results;
    \item The robustness of transferability on OOD downstream tasks constantly deteriorates as the noise increases, even with the improvement in ID tasks on $5\%$ noisy pre-trained models. 
\end{itemize}
\end{mdframed}

These observations seem to be in contrast to the common belief in noisy label learning, which primarily aims to correct/eliminate noise or perform robust learning against noise \cite{Ghosh2017RobustLF,Li2020DivideMixLW,sopliu22w,xue2022investigating}.
However, we show that the noise in pre-training can have certain benevolent effects on downstream tasks that is agnostic to model architectures, proxy pre-training objectives (FS and CLIP), and pre-training dataset scales (1M for ResNet-50 and ViT-B-16 FS, 15M for ResNet-50 CLIP, and 27M for ViT-B-16 CLIP). 

\subsubsection{Downstream results using LoRA}

On the basis of the LP results, a natural question arises: \emph{does the same observation hold for other tuning methods that actually modify the pre-trained parameters of these models}? 
Here, we try to answer this question utilizing the well-known parameter-efficient tuning method -- LoRA \cite{hu2021lora}, which partially modifies the pre-trained representations. 
LoRA and other parameter-efficient tuning methods \cite{he2021towards} have become common practice with the recent advances of large foundation models. 
On downstream tasks, LoRA introduces light-weight modules at each intermediate layer of transformers and trains only these modules with the backbone frozen. 

\textbf{Setup}. 
We implement LoRA on ViT-B-16 models from both FS and CLIP pre-training. 
Based on the PEFT of the Huggingface \cite{peft}, we insert LoRA modules at the fully connected layers of the MLP block at each layer of ViT-B-16, with a feature dimension reduction ratio of 8. 
We adopt the learning rate of 2$e$-4 and the weight decay of 1$e$-4 to train the LoRA modules in the ID and OOD downstream tasks with full data for three runs.

\begin{figure*}[t!]
\centering
    \hfill
    \subfloat[ResNet-50 FS, ID]{\label{fig:r50_in1k_id_tuning}\includegraphics[width=0.24\linewidth]{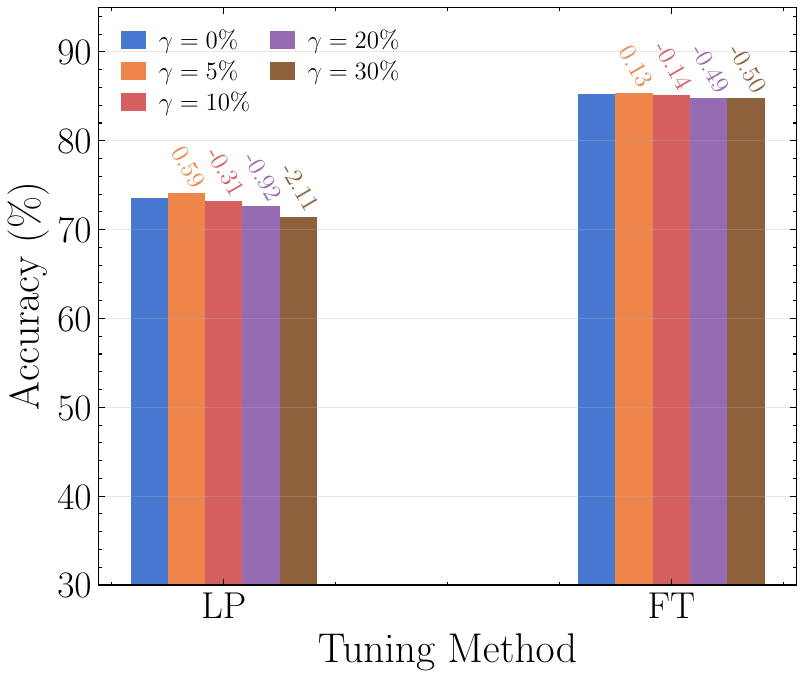}}
    \hfill
    \subfloat[ResNet-50 FS, OOD]{\label{fig:r50_in1k_odd_tuning}\includegraphics[width=0.24\linewidth]{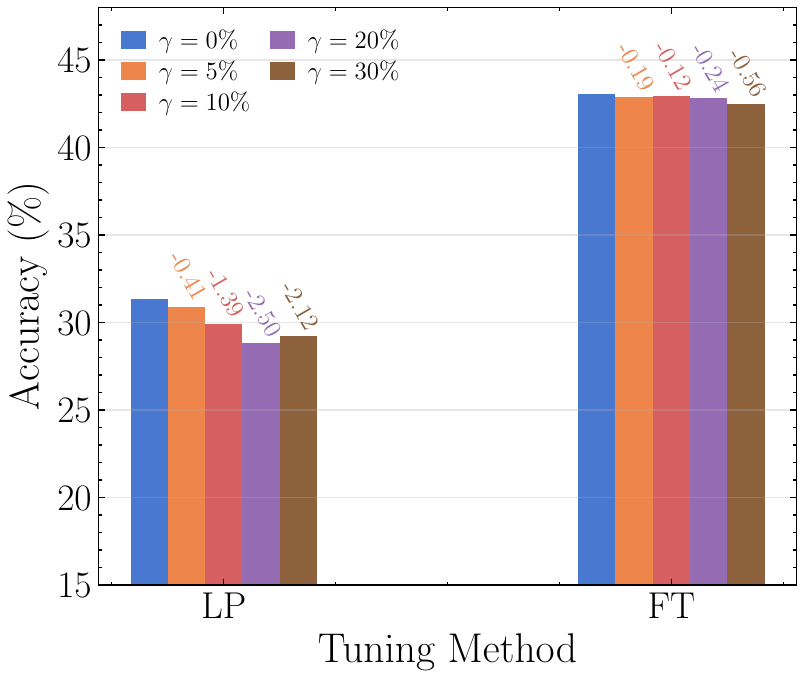}}
    \hfill
    \subfloat[ResNet-50 CLIP, ID]{\label{fig:r50_yfcc15m_id_tuning}\includegraphics[width=0.24\linewidth]{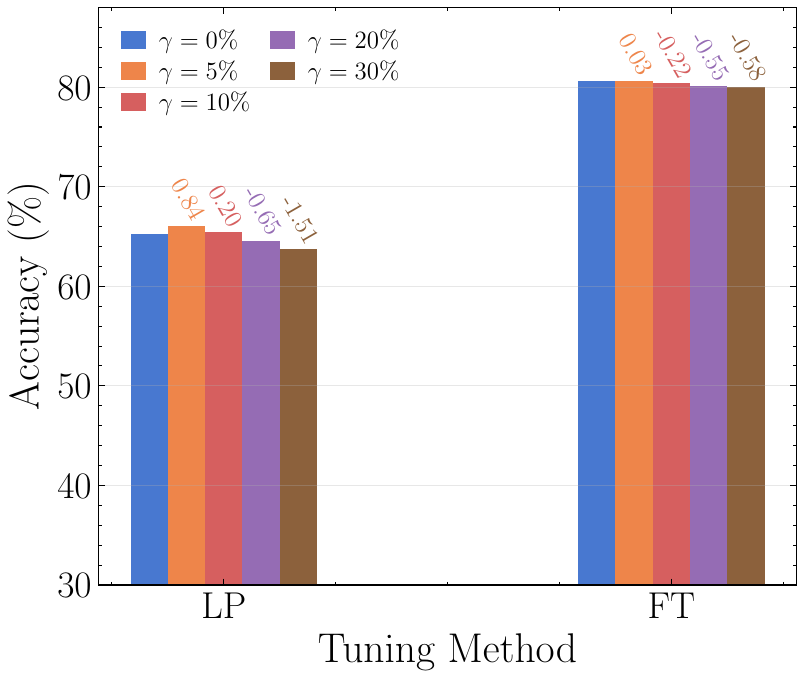}}
    \hfill
    \subfloat[ResNet-50 CLIP, OOD]{\label{fig:r50_yfcc15m_ood_tuning}\includegraphics[width=0.24\linewidth]{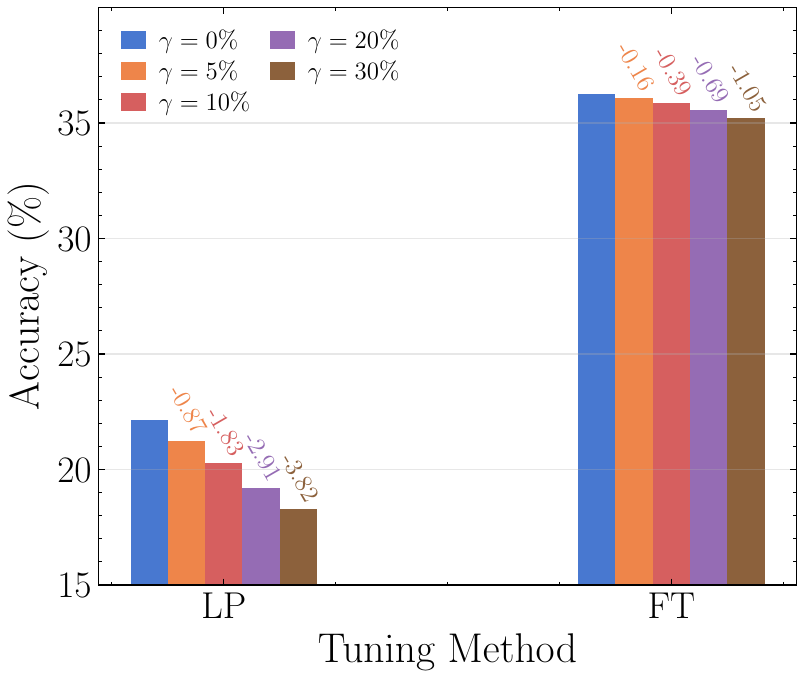}}
    \hfill

    \hfill
    \subfloat[ViT-B-16 FS, ID]{\label{fig:vitb16_in1k_id_tuning}\includegraphics[width=0.24\linewidth]{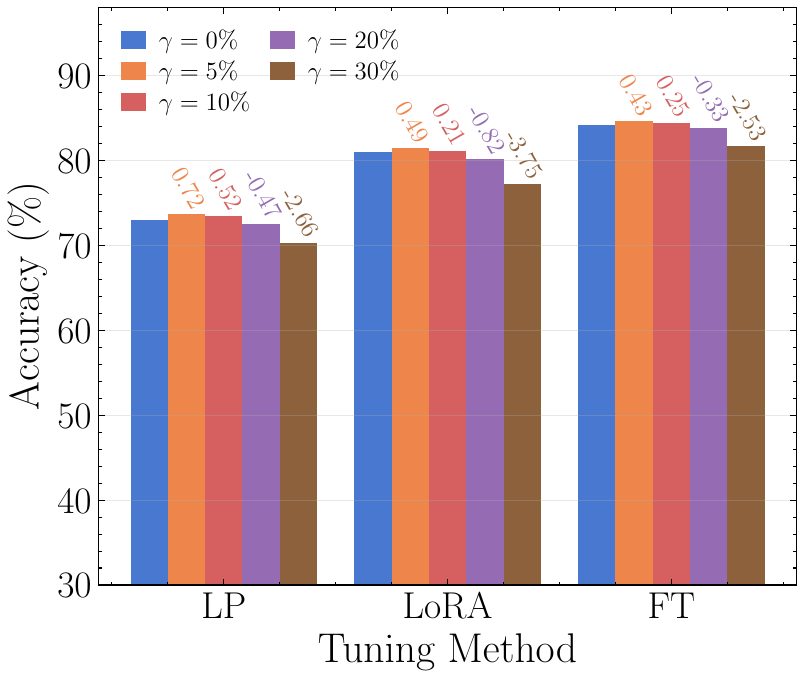}}
    \hfill
    \subfloat[ViT-B-16 FS, OOD]{\label{fig:vitb16_in1k_odd_tuning}\includegraphics[width=0.24\linewidth]{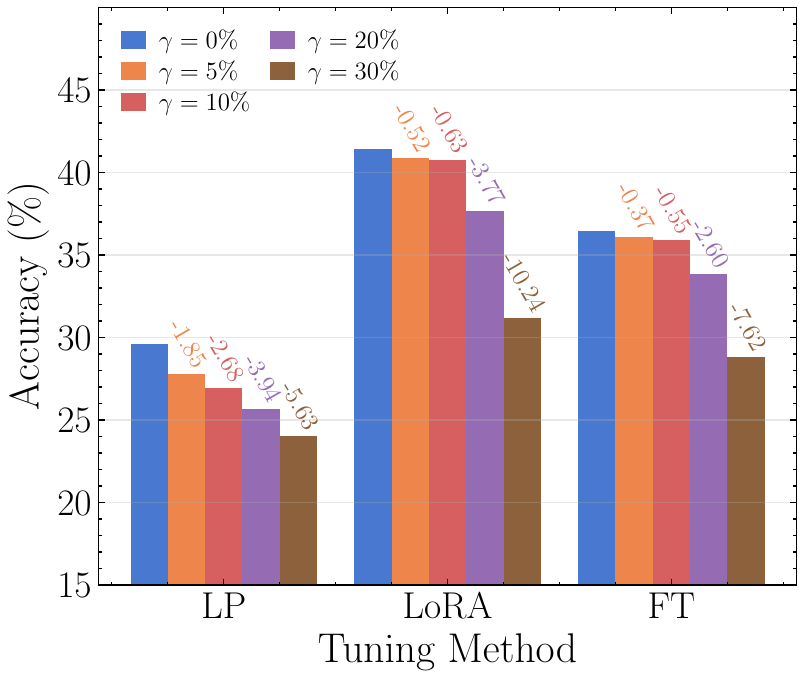}}
    \hfill
    \subfloat[ViT-B-16 CLIP, ID]{\label{fig:vitb16_yfcc15m_id_tuning}\includegraphics[width=0.24\linewidth]{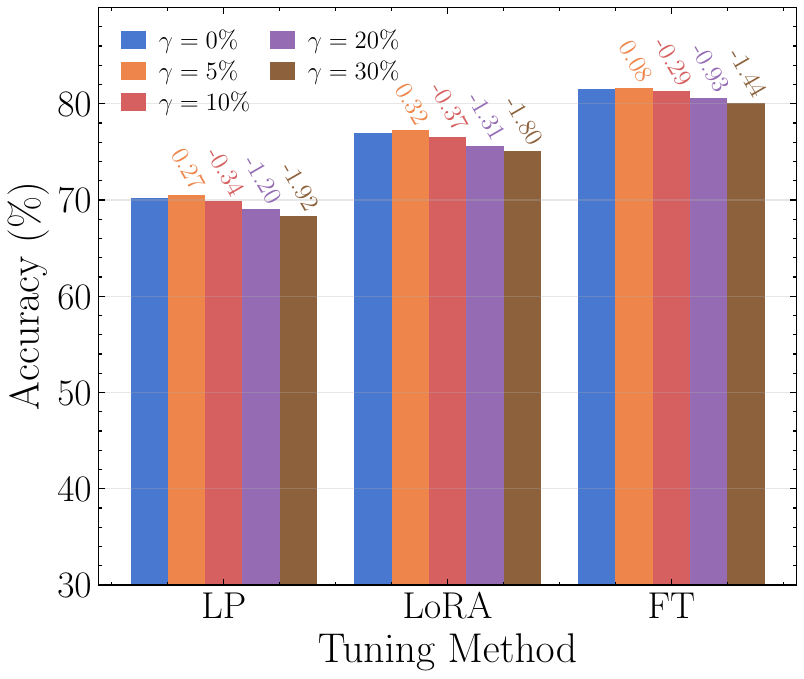}}
    \hfill
    \subfloat[ViT-B-16 CLIP, OOD]{\label{fig:vitb16_yfcc15m_ood_tuning}\includegraphics[width=0.24\linewidth]{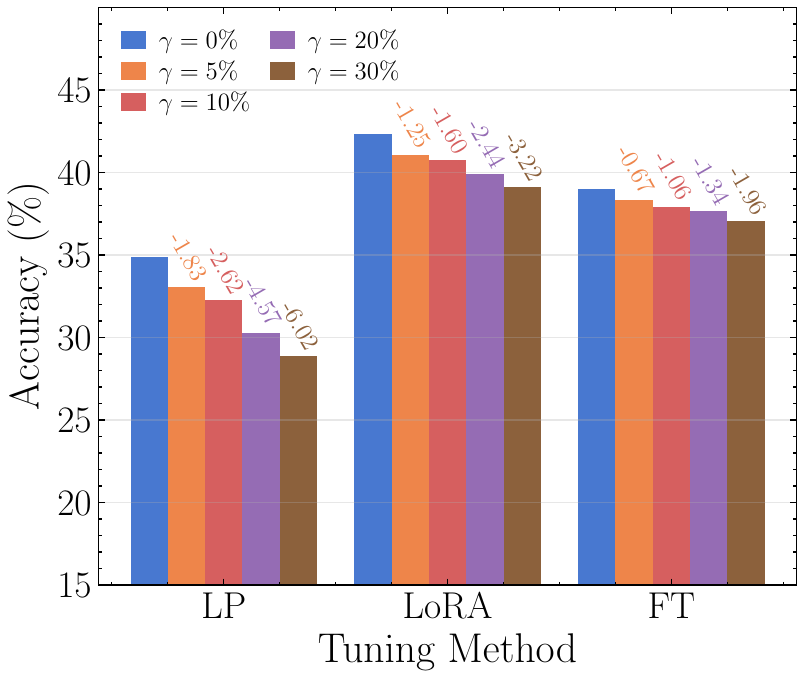}}
    \hfill

\caption{
Average ID and OOD tuning results of ResNet-50 (top row) and ViT-B-16 (bottom row), using ImageNet-1K (IN-1K) fully-supervised pre-training ((a), (b), (e), (f)) and YFCC15M (and CC12M) CLIP pre-training ((c), (d), (g), (h)) on downstream tasks with full data. 
For ResNet-50, we adopt linear probing (LP) and full fine-tuning (FT). 
For ViT-B-16, we additionally adopt LoRA. 
On different tuning methods, we find similar observations for downstream tasks, where slight noise in pre-training benefits the model's ID performance but always hurts the OOD performance. 
As more pre-trained parameters are modified on downstream tasks, i.e., from LP (to LoRA) to FT, the difference (shown on the top of each bar) between noisy pre-trained models becomes smaller in terms of both ID benefits (with slight noise) and OOD deterioration. 
}
\label{fig:r50_in_ood_tuning}
\vspace{-0.15in}
\end{figure*}

\textbf{Results}.
The LoRA results of ViT-B-16 on downstream ID and OOD tasks with full data are shown in \cref{fig:vitb16_in1k_id_tuning}-\cref{fig:vitb16_yfcc15m_ood_tuning}. 
From the results, one can observe that:
\begin{mdframed}
    \begin{itemize}[leftmargin=1em]
\setlength\itemsep{0em}
    \item LoRA tuning presents a similar trend on both ID and OOD tasks as in LP: slight noise in pre-training can benefit the ID while hurting the OOD performance;
    \item With pre-trained parameters and representations being modified by LoRA, the differences between the noisy and the clean pre-trained models, both improvement and deterioration, become smaller.
\end{itemize}
\end{mdframed}


\subsubsection{Downstream results using full fine-tuning}

We further conduct full fine-tuning (FT) experiments of both ResNet-50 and ViT-B-16 models, where all pre-trained parameters can be modified on downstream tasks. 
Although we can assume that the large foundation models are black-box due to their limited access, it would still be interesting to verify FT to fully answer the above question. 

\textbf{Setup}.
We evaluate the full FT also on ID and OOD tasks with full data. 
We set the learning rate and weight decay to 1$e$-4 and conduct FT for 30 epochs. 
Each experiment is run with three random seeds. 

\textbf{Results}.
The results of full FT are shown in \cref{fig:r50_in_ood_tuning} for both ResNet-50 and ViT-B-16, compared to LP (and LoRA for ViT-B-16). 
Interestingly, one can find:
\begin{mdframed}
    \begin{itemize}[leftmargin=1em]
\setlength\itemsep{0em}
    \item FT also present similar observations as LP and LoRA for both ID and OOD tasks, even though all pre-trained parameters can be changed;
    \item The differences between noisy and clean pre-trained model are further reduced with FT. Notably, the differences for the smaller model ResNet-50 are no longer prominent, yet for the larger model ViT-B-16, there is still a significant gap. 
\end{itemize}
\end{mdframed}


\subsection{Results on Detection and Segmentation Tasks}

In addition to classification, we also perform evaluations on detection and segmentation tasks using the popular IN-1K pre-trained ResNet-50 as the backbone.

\textbf{Setup}. 
The detection and segmentation evaluation is carried out on the COCO 2017 ID task \cite{lin2014microsoft}, which contains 118K training, 5K validation, and 20K test images. 
For the object detection task, we consider two common methods: Faster R-CNN \cite{ren2015faster} and RetinaNet \cite{lin2017focal}, where we utilize AdamW optimizer with a learning rate of 1$e$-4, weight decay of 0.01, and a batch size of 32. 
We train each method with the clean and noisy IN-1K pre-trained ResNet-50 for 24 epochs (2x schedule). 
For instance, for the segmentation task, we adopt Mask R-CNN \cite{he2017mask} and SOLOv2 \cite{wang2020solov2}.
For these two methods, we use SGD optimizer with a learning rate of 0.01, a weight decay of 1$e$-4, and a batch size of 32, and train them for 12 epochs (1x schedule).
During the adaptation of pre-trained models to detection and segmentation tasks, the deeper layers of the pre-trained ResNet-50 are usually frozen, and only the shallow layers and the modules related to detection and segmentation, such as FPN \cite{ren2015faster}, are trained.

\textbf{Results}. 
We present the results for object detection and instance segmentation of IN-1K FS pre-trained ResNet-50 on COCO in \cref{tab:r50_det}.
One can also observe that, for these tasks, slight pre-training noisy (up to 5\% or 10\%) pre-trained models can also benefit the downstream performance, achieving better mean average precision (mAP) at all scales or comparable to clean pre-trained models. 
Moreover, further increasing the noise ratio also leads to inferior performance.

\section{Feature Space Analysis of Noise}
\label{sec:analysis-empirical}

The results from previous section demonstrate that slight noise in pre-training can benefit the ID downstream tasks while always hurting the OOD tasks.
This observation is agnostic to model architectures, pre-training proxy objectives, pre-training data scales, tuning methods, and downstream applications. 
In this section, we attempt to answer the following fundamental question via the feature space analysis: \textit{where does the superior transferability (with slight noise) and the inferior robustness stem from}? 

\begin{figure*}[!t]
\centering
    \hfill
    \subfloat[ResNet-50 FS, ID]{\label{fig:r50_in1k_id_svd}\includegraphics[width=0.24\linewidth]{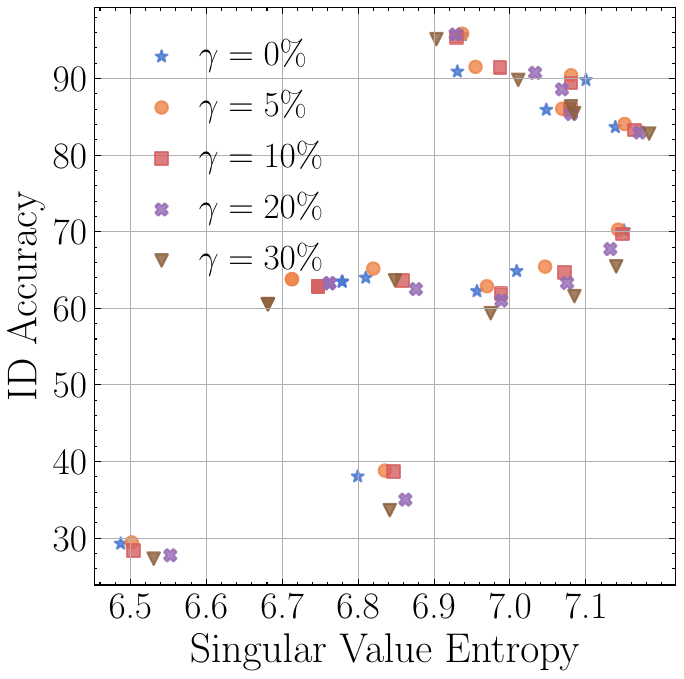}}
    \hfill
    \subfloat[ResNet-50 FS, OOD]{\label{fig:r50_in1k_ood_svd}\includegraphics[width=0.24\linewidth]{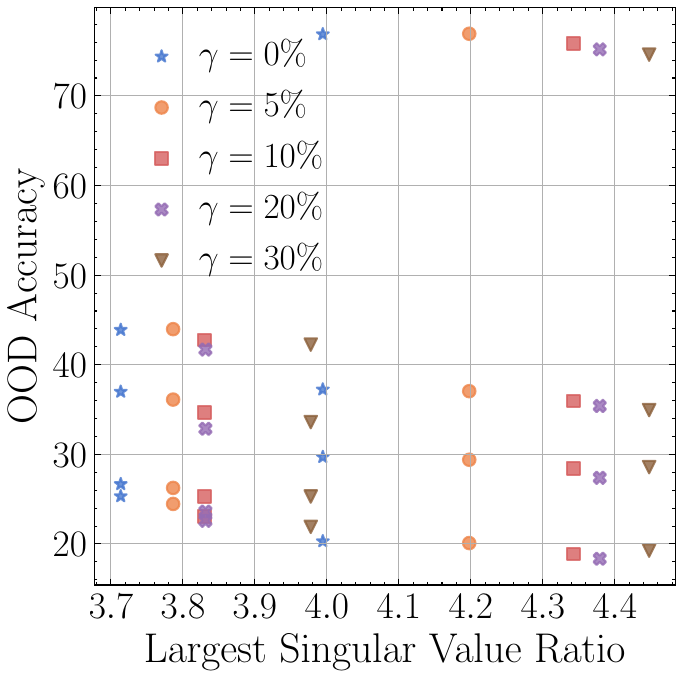}}
    \hfill
    \subfloat[ResNet-50 CLIP, ID]{\label{fig:r50_yfcc15m_in_svd}\includegraphics[width=0.24\linewidth]{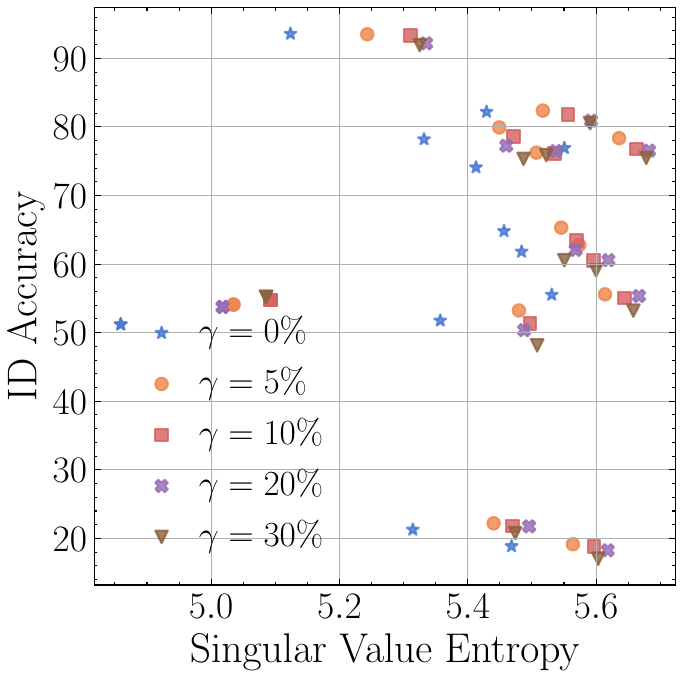}}
    \hfill
    \subfloat[ResNet-50 CLIP, OOD]{\label{fig:r50_yfcc15m_ood_svd}\includegraphics[width=0.24\linewidth]{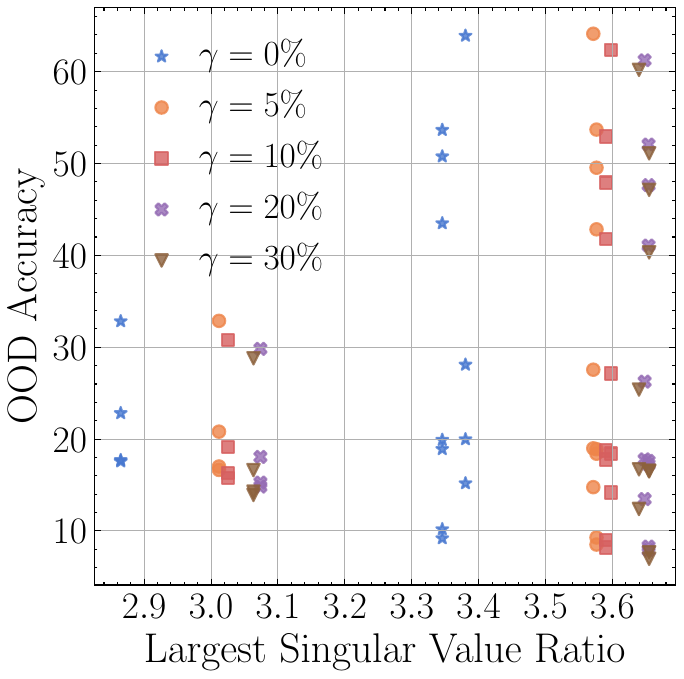}}
    \hfill

    \hfill
    \subfloat[ViT-B-16 FS, ID]{\label{fig:vitb16_in1k_id_svd}\includegraphics[width=0.24\linewidth]{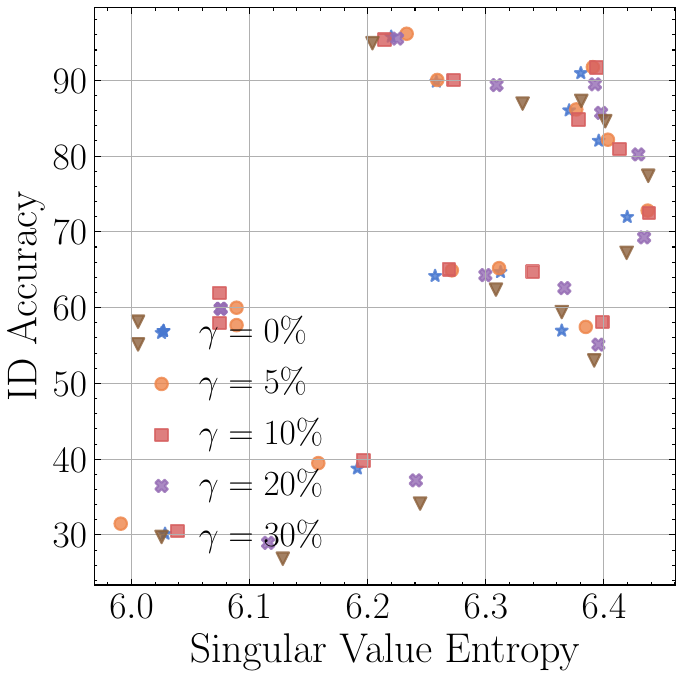}}
    \hfill
    \subfloat[ViT-B-16 FS, OOD]{\label{fig:vitb16_in1k_ood_svd}\includegraphics[width=0.24\linewidth]{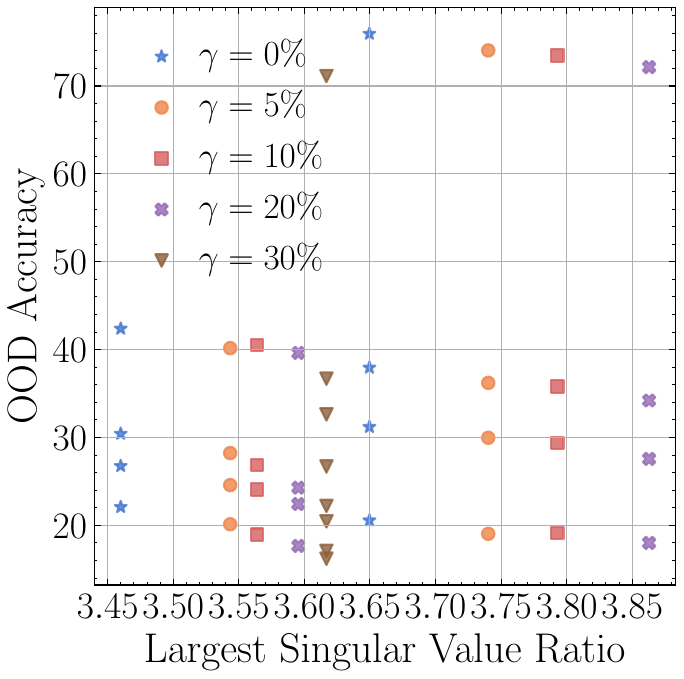}}
    \hfill
    \subfloat[ViT-B-16 CLIP, ID]{\label{fig:vitb16_yfcc15m_in_svd}\includegraphics[width=0.24\linewidth]{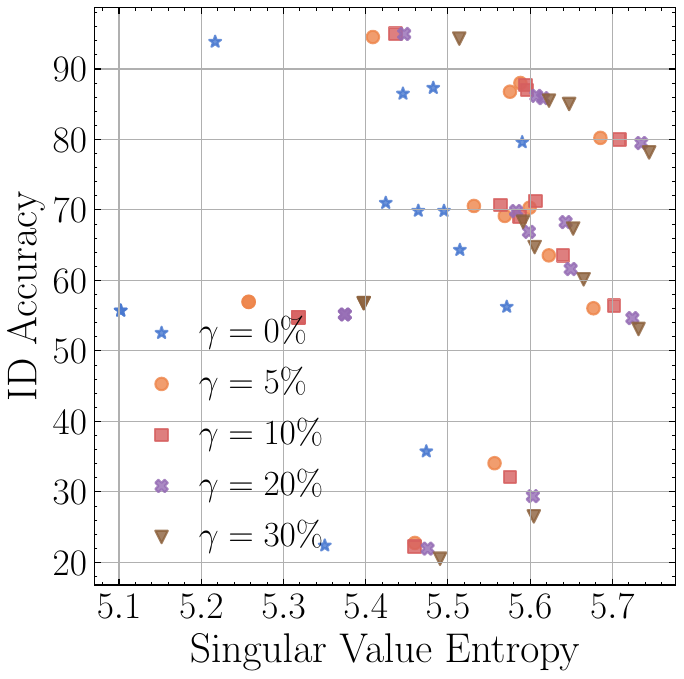}}
    \hfill
    \subfloat[ViT-B-16 CLIP, OOD]{\label{fig:vitb16_yfcc15m_ood_svd}\includegraphics[width=0.24\linewidth]{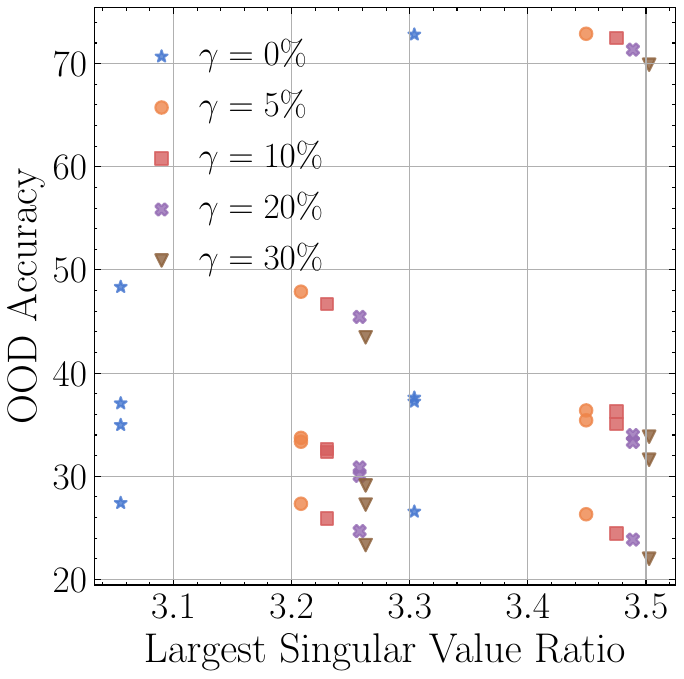}}
    \hfill

\caption{Feature SVD analysis of ResNet-50 (top row) and ViT-B-16 (bottom row). We compute the singular value entropy (SVE) for in-domain (ID) tasks and the largest singular value ratio (LSVR) for out-of-domain (OOD) tasks. Both metrics are computed for ImageNet-1K fully-supervised pre-trained ((a), (b), (e), (f)) and YFCC15M (and CC12M) CLIP pre-trained ((c), (d), (g), (h)) models. For both model architectures, the SVE first slightly improves as the noise ratio increases to $5\%$ or $10\%$, indicating better generalization. As the noise ratio increases, the SVE further improves, and the LSVR drops significantly, corresponding to worse generalization on OOD tasks. The dominant singular components become less transferable. 
} 
\label{fig:r50_id_ood_svd}
\vspace{-0.15in}
\end{figure*}

To understand the noise in pre-training data, we empirically analyze the singular value spectrum of the pre-trained feature space on downstream classification datasets (ID and OOD), which is widely considered to be related to the generalization performance \cite{oymak2019generalization, chen19itrans, xue2022investigating}.
More specifically, we perform singular value decomposition (SVD) on the features $\mathbf{F} \in \mathbb{R}^{M \times D}$ of pre-trained feature extractors on a downstream dataset:
$\mathbf{F} =\mathbf{U} \boldsymbol{\Sigma} \mathbf{V}^{\top}$, where we denote $M$ as the number of samples in downstream datasets and $D$ as the feature dimension. and assume $D \leq M$ \cite{kumar2022fine}. 
$\mathbf{U}$ and $\mathbf{V}$ denote the left and right singular vector matrices, respectively, and $\mathbf{\Sigma}$ denote the diagonal singular value matrix $\{\sigma_1, \ldots, \sigma_D\}$.
Based on the singular values, we define two metrics that can help to understand the observations:
\begin{definition}[Singular Value Entropy] 
The singular value entropy (SVE) is defined as the entropy of normalized singular values, which measures the flatness of the singular value distribution of the pre-trained models:
\begin{equation}
     \mathrm{SVE} = -\sum_{i=1}^D \frac{\sigma_i}{\sum^{D}_{j=1} \sigma_j} \log \frac{ \sigma_i}{\sum^{D}_{j=1} \sigma_j}.
\end{equation}
Larger SVE values indicate that the feature space captures more structure in the data and thus spans more dimensions either due to more discriminated features are learned or memorization of the noise structure during pre-training. 
\end{definition}

\begin{table}[t!]
\centering
\caption{Object detection and instance segmentation results on COCO 2017 of IN-1K ResNet-50 noisy FS pre-trained models.}
\label{tab:r50_det}
\vspace{-0.1in}
\resizebox{0.98 \columnwidth}{!}{%
\begin{tabular}{@{}c|c|ccc|c|ccc@{}}
\toprule
\multirow{2}{*}{ Noise (\%) } & \multirow{2}{*}{ Method } & \multicolumn{3}{c|}{Object Detection} & \multirow{2}{*}{ Method } & \multicolumn{3}{c}{Instance Segmentation} \\ 
 &  & \multicolumn{1}{c}{$\textrm{AP}^{\textrm{box}}$} & \multicolumn{1}{c}{$\textrm{AP}^{\textrm{box}}_{50}$} & \multicolumn{1}{c|}{$\textrm{AP}^{\textrm{box}}_{75}$}  & & \multicolumn{1}{c}{$\textrm{AP}^{\textrm{mask}}$} & \multicolumn{1}{c}{$\textrm{AP}^{\textrm{mask}}_{50}$} & \multicolumn{1}{c}{$\textrm{AP}^{\textrm{mask}}_{75}$} \\ \midrule
0 & \multirow{5}{*}{Faster R-CNN \cite{ren2015faster}}   & 38.5 & 59.8 & 41.7  &  \multirow{5}{*}{Mask R-CNN \cite{he2017mask}}  & 31.3 & \textbf{51.3} & 33.0 \\
                              5 &  & \textbf{38.6} & \textbf{60.1} & \textbf{41.9}   &   & \textbf{31.4} & \textbf{51.3} & \textbf{33.2} \\
                              10 &  & \textbf{38.6} & 60.0   & \textbf{41.9} &   & \textbf{31.3} & \textbf{51.3}   & 32.9 \\
                              20 &  & 38.4 & 59.7 & 41.6 &   & 31.2 & 51.1 & 32.8  \\
                              30  & & 37.9 & 59.1 & 40.9 &   & 30.3 & 49.9 & 32.1  \\ \midrule
 0 & \multirow{5}{*}{RetinaNet \cite{lin2017focal}}    & 38.3 & 58.2 & 40.9  &  \multirow{5}{*}{SOLOv2 \cite{wang2020solov2}}  & 32.2 & 52.7 & 33.6\\
                              5  &  & \textbf{38.4} & \textbf{58.4} & 40.9 &   & \textbf{32.7} & \textbf{53.2} & \textbf{34.2}  \\
                              10  &  & \textbf{38.4 }& 58.1 & \textbf{41.1}    & & 32.4 & 52.8 & 33.9  \\
                              20  & & 37.9 & 57.7 & 40.4 &   & 32.0 & 52.2 & 33.6 \\
                              30 &  & 37.0   & 56.8 & 39.1 &   & 31.4   & 51.3 & 32.5 \\ \bottomrule
\end{tabular}%
}
\vspace{-0.2in}
\end{table}

\begin{definition}[Largest Singular Value Ratio] 
The largest singular value ratio (LSVR) is defined as the negative logarithm of the ratio of the largest singular value $\sigma_1$ to the summation of all singular values of the pre-trained models:
\begin{equation}
    \mathrm{LSVR} = -\log \frac{\sigma_1}{\sum^{D}_{i=1} \sigma_i}.
\end{equation}
LSVR measures the variations in data captured by the singular vector corresponding to the largest singular value $\sigma_1$, which is related to the transferability of a model \cite{chen19itrans}.
\end{definition}

\textbf{Analysis.}
We plot the SVE for ID tasks and the LSVR for OOD tasks, and connect them to the performance of the LP, as shown in \cref{fig:r50_id_ood_svd}. 
Here, we consider LP only for simplicity and better visualization purpose. 
The performance of LoRA and FT can be represented by LP since they present the same trend. 
For ID tasks, as the noise ratio slightly increases, the learned representation usually presents slightly higher SVE, which indicates the pre-trained feature extractor captures more structure in data. 
Specifically, more capabilities of the feature space are assigned to fit the noise in data, resulting in a feature space spanning more dimensions, 
which provides better-initialized features on downstream tasks and facilitates generalization. 
Similar observations have also been found and explored in \cite{wu2022noisytune}.
However, as the noise ratio further increases, the increased SVE indicates that a more noisy data structure is captured and memorized, thus leading to deteriorated generalization performance. 
When the labels in the pre-training are random, the SVE of the feature extractor would further increase by memorizing all the noise but would not generalize to downstream tasks, similar to \cite{zhang2021understanding}. 
For OOD tasks, the robustness performance is \emph{negatively correlated} with the LSVR. 
As the noise ratio increases, the LSVR increases consistently with the decreasing largest singular value. 
A less transferable component is learned, thus resulting in a poorer generalization on unseen OOD tasks.
For both ResNet-50 and ViT-B-16 and FS and CLIP pre-training objectives, the trends of the SVE change in ID tasks and the LSVR change in OOD tasks are similar, showing a strong correlation of previously observed phenomenon. 

\section{Mitigating the Pre-training Noise}
\label{sec:mitigate}

\begin{figure}
\centering
    \subfloat[Black-box NMTune]{\label{fig:mlp-arch}\includegraphics[width=0.45\linewidth]{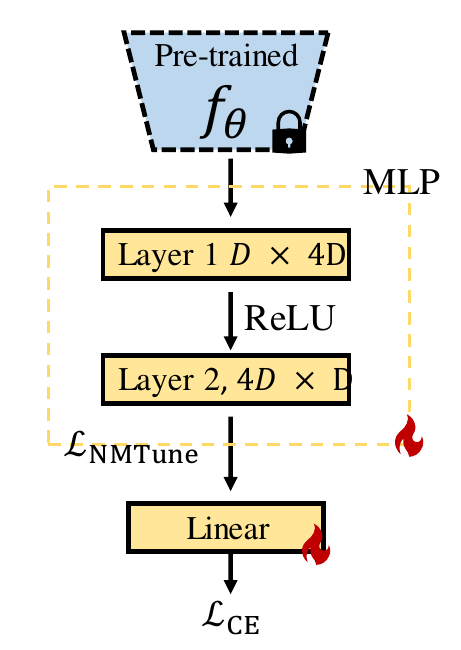}}
    \hfill
    \subfloat[Param.-Efficient NMTune]{\label{fig:lora-arch}\includegraphics[width=0.45\linewidth]{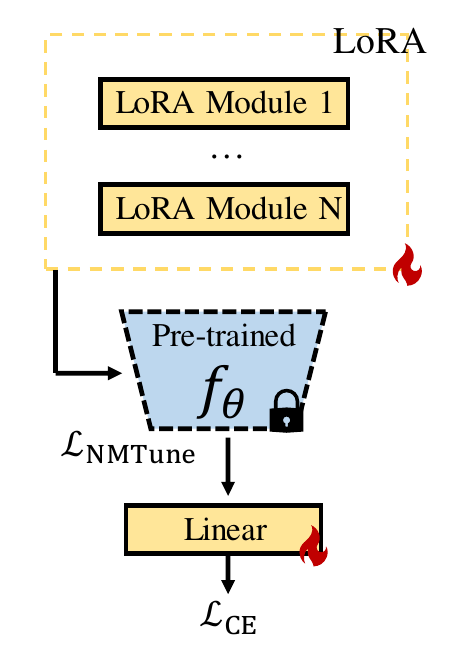}}
\caption{Illustration of NMTune in (a) black-box tuning, i.e., LP and (b) parameter-efficient tuning, i.e., LoRA.}
\label{fig:nmtune-mlp-lora}
\vspace{-0.2in}
\end{figure}

In this section, we propose a tuning method with regularization on downstream tasks based on the above analysis, which we call ``Noisy Model Tuning'' (NMTune, \cref{fig:method} and \ref{fig:nmtune-mlp-lora}) in response to the noisy model learning setting. 
We demonstrate that NMTune can boost the generalization performance and rectify the generalization behaviors affected by the pre-training noise on downstream tasks.

\subsection{Method}
\label{sec:mitigate-method}

Per the analysis above, noise in pre-training can shape the feature space differently from pre-training on clean data, reducing the top dominant singular values with dampened transferability while increasing the spanning dimensions of the feature space to fit the noise structure.
Considering the large pre-trained models are usually difficult to fully fine-tune due to the enormous parameter size and limited computation resources, we propose to alter the pre-trained feature space $\mathcal{F}$ in both the light-weight parameter-efficient and black-box paradigms, i.e., LoRA and LP. 
More specifically, for LP, we introduce a multi-layer perceptron (MLP) $h_{\omega}$ transforming the pre-trained features into new feature space $\mathcal{Z}$. 
For LoRA, since it inserts learnable models at intermediate layers of the pre-trained model, it is already capable to reshape the pre-trained features, and thus we do not utilize additional MLP on top of the pre-trained models. 
The new feature space of LoRA is also denoted as $\mathcal{Z}$. 
We propose three straightforward regularization terms on $\mathbf{Z}$ to encourage the pre-trained knowledge to be maintained and directly improve SVE and LSVR of the new feature space. 

\textbf{Consistency regularization}. 
To encourage the consistency of the pre-trained knowledge for the new feature space, we adopt a mean-square-error (MSE) loss between the normalized features $\mathbf{F}$ and $\mathbf{Z}$:
\begin{equation}
    \mathcal{L}_{\mathrm{MSE}} = \left\| \frac{\mathbf{F}}{\| \mathbf{F} \|_2}  - \frac{\mathbf{Z}}{\| \mathbf{Z} \|_2} \right\|_2^2.
\end{equation}
This objective facilitates inheriting the pre-trained knowledge in the transformed features $\mathbf{Z}$. 

\textbf{Covariance regularization.}
We define the covariance loss to encourage the off-diagonal elements in the covariance matrix of the transformed feature $C(\mathbf{Z})$ to be close to $\mathbf{0}$:
\begin{equation}
\begin{split}
    \mathcal{L}_{\mathrm{COV}} &= \frac{1}{D} \sum_{i \neq j}[C(\mathbf{Z})]_{i, j}^2, \\
    \text{where } C(\mathbf{Z}) &= \frac{1}{M-1} \sum_{i=1}^M\left(z_i-\bar{z}\right)\left(z_i-\bar{z}\right)^T, \\
    \text{and } \bar{z}&=\frac{1}{M} \sum_{i=1}^M z_i.
\end{split}
\end{equation}
Inspired by Zbontar et al. \cite{zbontar2021barlow} and Bardes et al. \cite{bardes2021vicreg}, we use the covariance regularization term to improve the SVE of feature space by preventing the different coordinates of the features from encoding similar information. 
It also encourages more discriminative features to be learned.

\textbf{Dominant singular value regularization}.
To help transferability, we use a specific regularization to improve the LSVR by maximizing the ratio of the largest singular value:
\begin{equation}
    \mathcal{L}_{\mathrm{SVD}} = - \frac{\sigma_{1}}{\sum_{j=1}^{D} \sigma_{j}}.
\end{equation}

\noindent The total objective on a downstream task thus becomes:
\begin{equation}
\begin{split}
       \mathcal{L} &= \mathcal{L}_{\mathrm{CE}} + \lambda \mathcal{L}_{\mathrm{NMTune}} \\
       & = \mathcal{L}_{\mathrm{CE}} + \lambda \left(  \mathcal{L}_{\mathrm{MSE}} + \mathcal{L}_{\mathrm{COV}} + \mathcal{L}_{\mathrm{SVD}} \right), 
\end{split}
\end{equation}
where $\mathcal{L}_{\text{CE}}$ is the cross-entropy loss for downstream classification.
We set $\lambda=0.01$ for all experiments in the following sections.

\subsection{Evaluation of NMTune}

Here, we evaluate the proposed NMTune on the noisy models and analyze the reason for its effectiveness.
For the black-box paradigm, we compare against solely training the MLP without the regularization, termed as MLP tuning, to show the effectiveness stems from the regularization rather than the extra parameters.
For the light-weight paradigm, we compare against LoRA without the regularization terms.

\begin{figure*}[!t]
\centering
    \hfill
    \subfloat[ResNet-50 ID F1]{\label{fig:r50_results_id_f1}\includegraphics[width=0.24\linewidth]{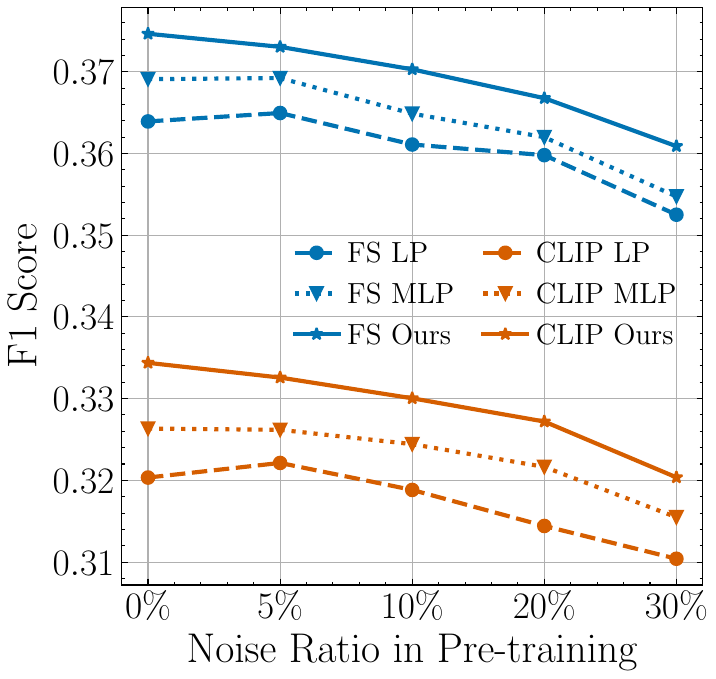}}
    \hfill
    \subfloat[ResNet-50 ID SVE]{\label{fig:r50_results_id_sve}\includegraphics[width=0.24\linewidth]{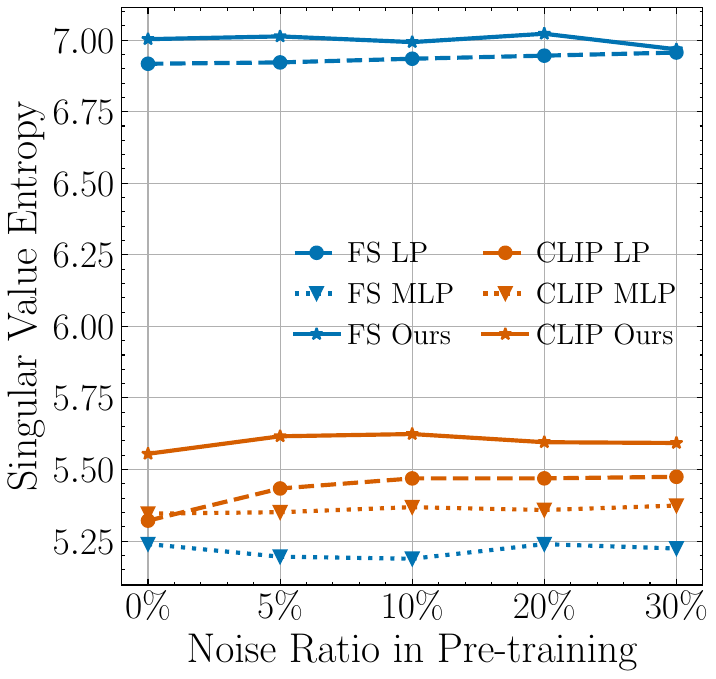}}
    \hfill
    \subfloat[ResNet-50 OOD F1]{\label{fig:r50_results_ood_f1}\includegraphics[width=0.24\linewidth]{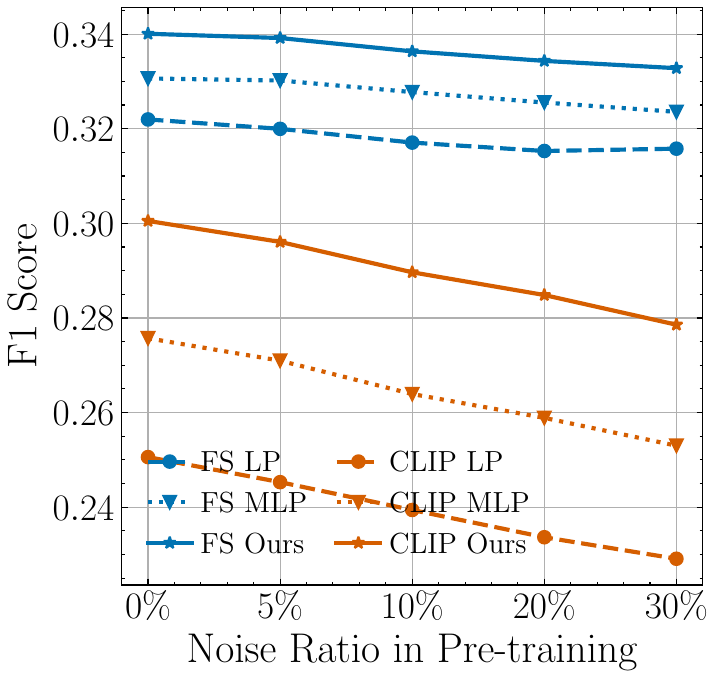}}
    \hfill
    \subfloat[ResNet-50 OOD LSVR]{\label{fig:r50_results_ood_lsvr}\includegraphics[width=0.24\linewidth]{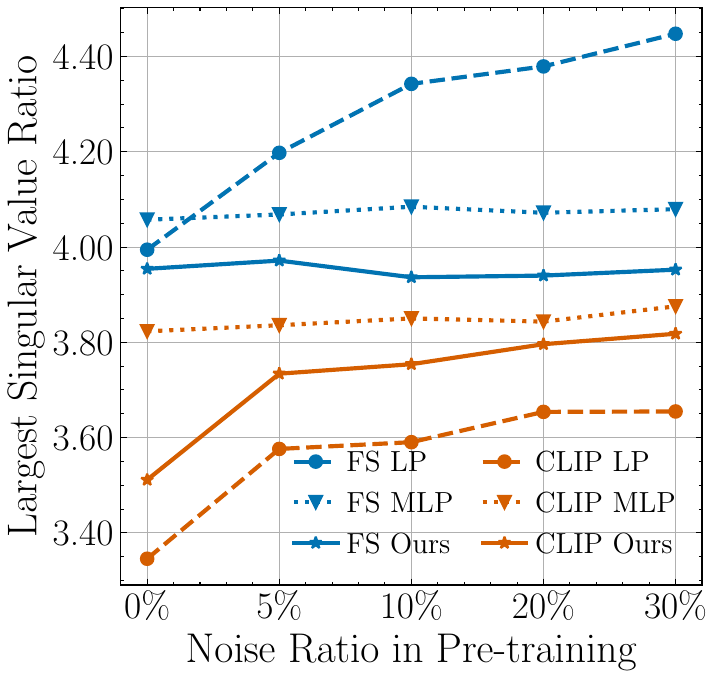}}
    \hfill

    \hfill
    \subfloat[ViT-B-16 ID F1]{\label{fig:vitb16_results_id_f1}\includegraphics[width=0.24\linewidth]{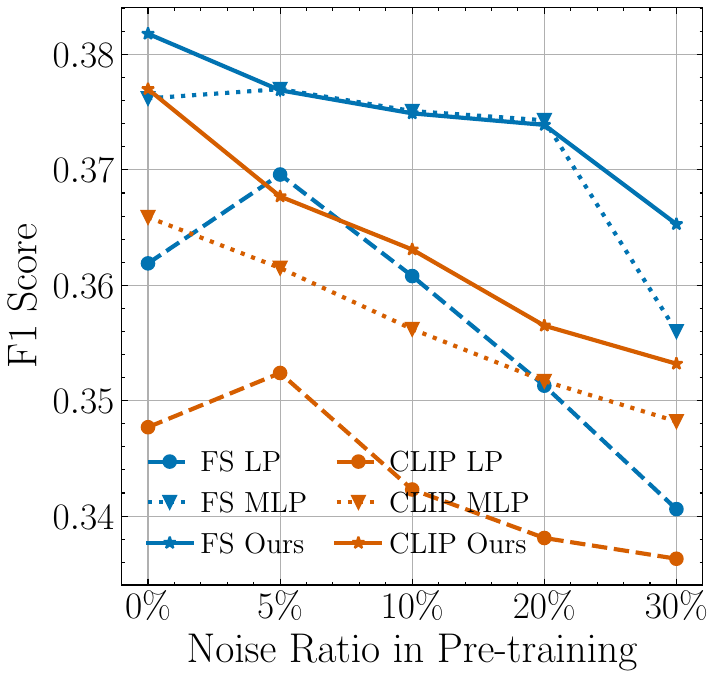}}
    \hfill
    \subfloat[ViT-B-16 ID SVE]{\label{fig:vitb16_results_id_sve}\includegraphics[width=0.24\linewidth]{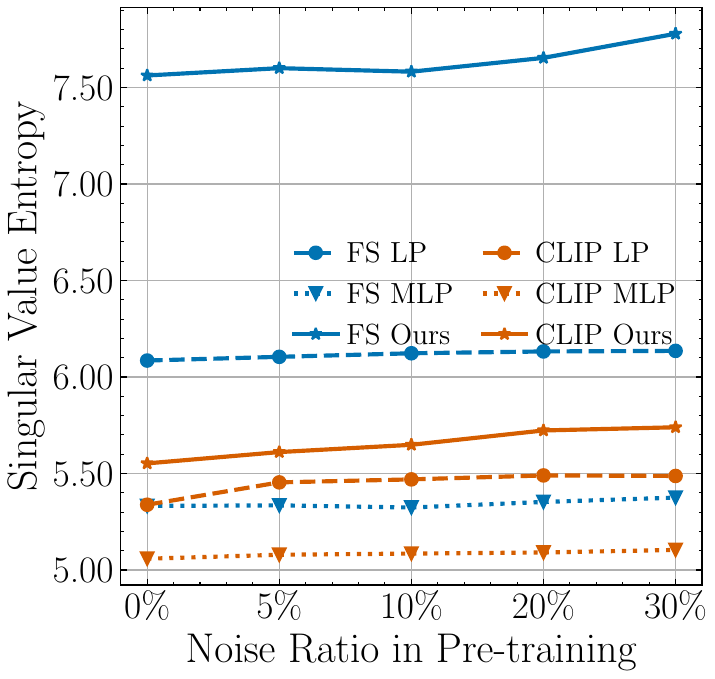}}
    \hfill
    \subfloat[ViT-B-16 OOD F1]{\label{fig:vitb16_results_ood_f1}\includegraphics[width=0.24\linewidth]{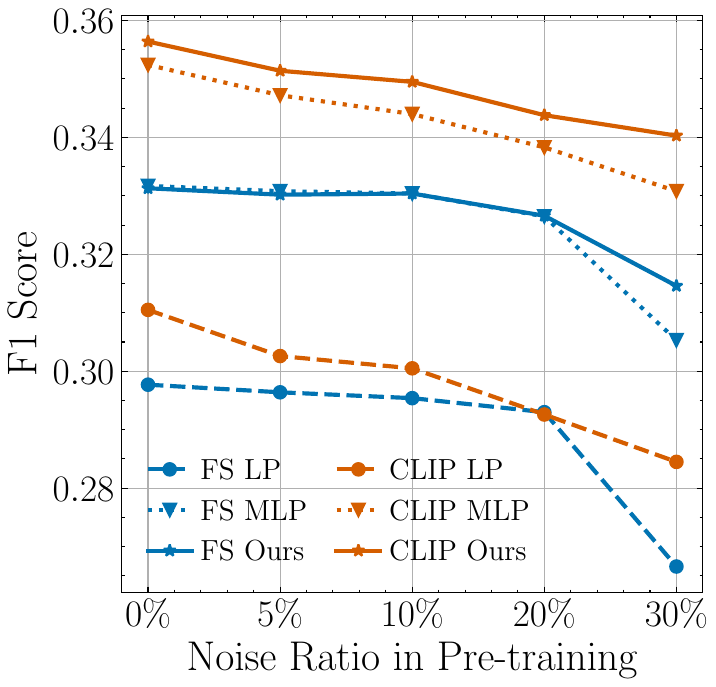}}
    \hfill
    \subfloat[ViT-B-16 OOD LSVR]{\label{fig:vitb16_results_ood_lsvr}\includegraphics[width=0.24\linewidth]{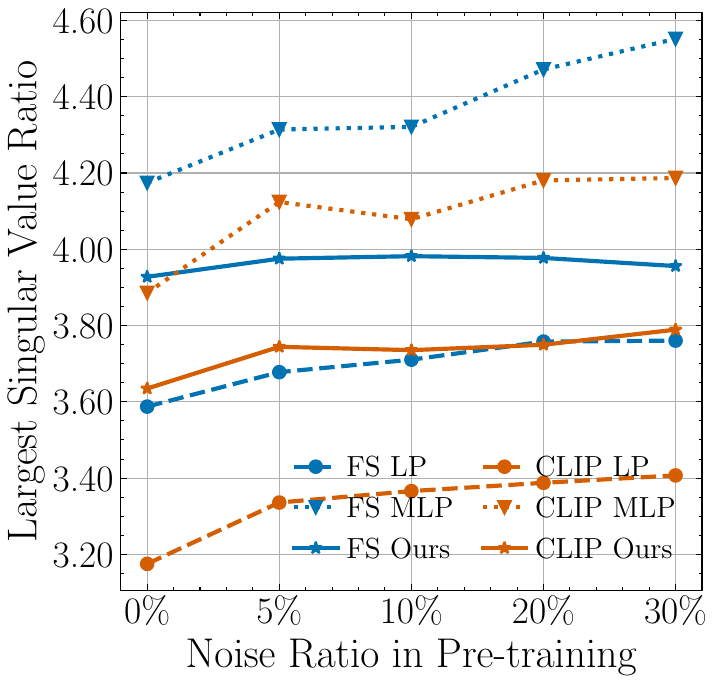}}
    \hfill

\caption{
Evaluation of our method (NMTune) in a black-box manner on ID and OOD downstream tasks, compared to MLP tuning and LP of ResNet-50 (top row) and ViT-B-16 (bottom row) FS pre-trained on ImageNet-1K (IN-1K) and CLIP pre-trained YFCC15M (and CC12M). (a) Average F1 score of ResNet-50 on ID tasks; (b) SVE of ResNet-50 on ID tasks; (c) Average F1 of ResNet-50 score on OOD tasks; (d) LSVR of ResNet-50 on OOD tasks; (e) F1 of ViT-B-16 on ID tasks; (f) SVE of ViT-B-16 on ID tasks; (g) F1 of ViT-B-16 on OOD tasks; (h) LSVR of ViT-B-16 on OOD tasks. Our method presents better SVE and LSVR on both ID and OOD tasks with better generalization performance. Our method also rectifies the malignant noise effect: the feature extractor pre-trained on clean data now exhibits better performance than others on noisy data on ID tasks; and the performance gap between the clean one and the noisy ones becomes smaller on OOD tasks.} 
\label{fig:r50_results}
\vspace{-0.2in}
\end{figure*}

\textbf{Black-box NMTune with MLP}. 
For ID tasks, we plot the average F1 score and SVE in \cref{fig:r50_results_id_f1} and \cref{fig:r50_results_id_sve} for ResNet-50 and \cref{fig:vitb16_results_id_f1} and \cref{fig:vitb16_results_id_sve} for ViT-B-16, respectively. 
The F1 score of linear probing (LP) on different pre-training noise ratios follows the same trend as the accuracy: it first increases as the noise ratio goes up to $5\%$ and then decreases. 
While adding an MLP can improve the F1 score in general, we find that it cannot mitigate the effect of noise, i.e., the clean pre-trained model still underperforms the $5\%$ noisy pre-trained models.
Further introducing our method can rectify the effect of noise on ID tasks, leading the clean pre-trained feature extractor to achieve the best results for both models and pre-training paradigms. 
More interestingly, only adding an MLP to LP can result in a smaller SVE, especially on ImageNet-1K, corresponding to a much sparser feature structure.
In contrast, our method provides a larger and flatter SVE, compared to LP.
It indicates the transformed feature space not only maintains the pre-trained knowledge but also spans more dimensions.
For OOD tasks, the F1 score and LSVR are shown in \cref{fig:r50_results_ood_f1} and \ref{fig:r50_results_ood_lsvr} for ResNet-50 and \cref{fig:vitb16_results_ood_f1} and \cref{fig:vitb16_results_ood_lsvr} for ViT-B-16, respectively. 
Similarly, one can observe significantly better generalization performance deploying NMTune, compared to the MLP and LP. 
We also notice a smaller performance gap for the clean pre-trained feature extractor and $5\%$ noisy pre-trained, especially for CLIP pre-training.
On LSVR, MLP tuning usually imposes larger LSVR compared to LP, presenting smaller dominant singular values. 
Considering MLP tuning also presents smaller SVE, its resulting feature space is expected to present a more long-tailed spectrum than the original feature space.
Maximizing the dominant singular values results in better transferability for OOD tasks, corresponding to the smaller relative values of the LSVR in noisy pre-trained models for NMTune.

\begin{figure}[!t]
\centering
    \subfloat[ViT-B-16 ID]{\label{fig:vitb16_lora_id_results}\includegraphics[width=0.48\linewidth]{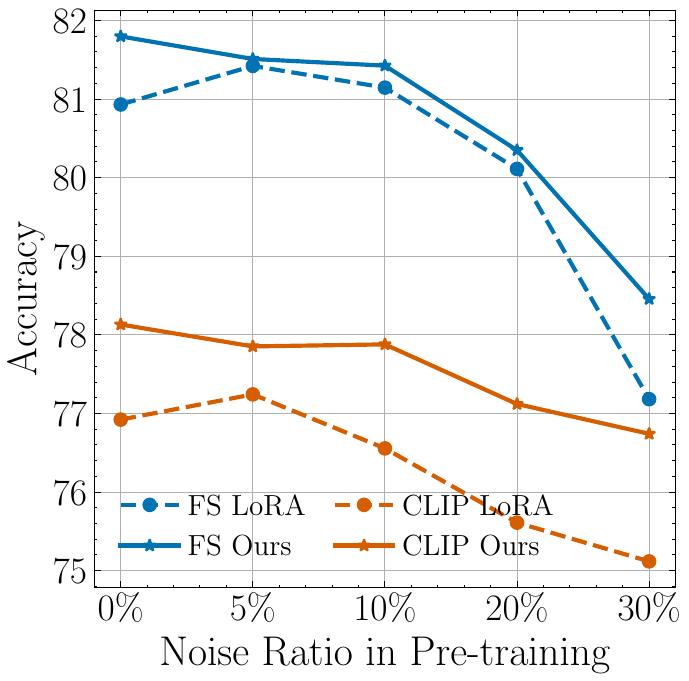}}
    \hfill
    \subfloat[ViT-B-16 OOD]{\label{fig:vitb16_lora_odd_results}\includegraphics[width=0.48\linewidth]{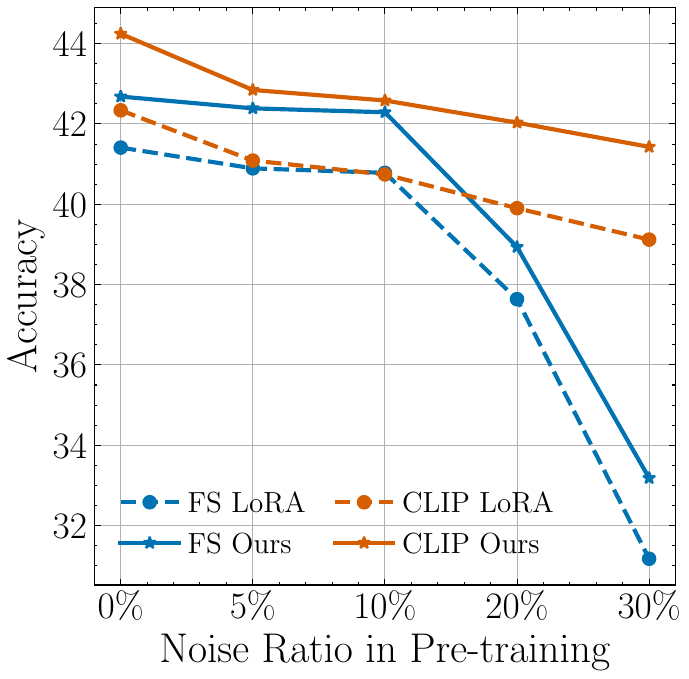}}
\caption{
Evaluation of our method (NMTune) in light-weight manner with LoRA on ID and OOD downstream tasks, compared to vanilla LoRA ViT-B-16 FS pre-trained on ImageNet-1K (IN-1K) and CLIP YFCC15M and CC12M pre-trained. (a) Average accuracy on ID tasks; (b) Average accuracy on OOD tasks. Our method presents better performance with smaller difference between the clean and noisy pre-trained models.} 
\label{fig:vitb16_lora_results}
\vspace{-0.1in}
\end{figure}

\textbf{Light-weight NMTune with LoRA}. 
We apply LoRA and LoRA with NMTune to ViT-B-16 pre-trained models. 
Compared to black-box tuning, LoRA with NMTune adjusts the pre-trained feature space directly with only a small proportion of parameters introduced.
For simplicity, we compare the accuracy of LoRA and LoRA with NMTune on ID and OOD tasks, as shown in \cref{fig:vitb16_lora_results}.
Parameter-efficient NMTune can rectify the behavior of noisy pre-trained models, whereas LoRA alone cannot. 
Similarly, NMTune also improves the generalization performance in general on both ID and OOD downstream tasks.

\section{Experiments on Practical Scenarios}
\label{sec:exp}

The main purpose of the simulated experiments in previous sections is to ease the analysis on the impact of pre-training noise on downstream tasks.
In this section, we turn to practical scenarios with various datasets and models to show the effectiveness of the proposed method.\footnote{Note that in these practical settings, it is infeasible to perform fine-grained analysis as previous sections due to the large volume or black-box nature of the pre-training data. Thus, we can only evaluate the efficacy of the mitigation algorithm.}
We adopt large-scale vision and language models that are commonly believed to be pre-trained on noisy data.\footnote{In fact, most of the popular pre-training datasets contain noise that cannot be perfectly cleaned due to their huge volume.}
Our evaluation results demonstrate that NMTune can boost the generalization performance of different noisy pre-trained models. 
We also provide discussions on asymmetric and random pre-training noise, the combination of noisy model \revision{transfer} learning and noisy label learning, running time analysis of NMTune in this section.

\subsection{Vision Models and Datasets}

\begin{table}[t]
\centering
\caption{Noisy vision and language models details.}
\label{tab:vision-model-details}
\vspace{-0.15in}
\resizebox{1.0 \linewidth}{!}{%
\begin{tabular}{@{}cccc@{}}
\toprule
Model & Pre-trained Data & Pre-trained Method & Param. Size (M) \\ \midrule
EfficientNet-B3 \cite{tan2019efficientnet}      &    JFT-300M \cite{hinton2015distilling}             &      Noisy Student \cite{xie2020ns}          &          12      \\ \midrule
ResNetv2-152x2 \cite{he2016identity}      &          ImageNet-21K   \cite{ridnik2021imagenet21k}     &                  BiT \cite{kolesnikov2020big}    &        236  \\ \midrule
Swin-L \cite{liu2021swin}      &          ImageNet-21K   \cite{ridnik2021imagenet21k}         &           Supervised \cite{liu2021swin}         &        197    \\ \midrule
ViT-L \cite{dosovitskiy2020image}      &         Laion-2B    \cite{schuhmann2022laionb}     &         CLIP  \cite{radford2021learning}            &    304     \\ \midrule 
ConvNext-L \cite{liu2022convnext}      &            Laion-2B   \cite{schuhmann2022laionb}         &           CLIP  \cite{radford2021learning}           &   200     \\ \midrule
BERT-L \cite{devlin2018bert}      &       \begin{tabular}[c]{@{}c@{}}BooksCorpus and \\ English Wikipedia\end{tabular}     &      Masked Modeling \cite{devlin2018bert}        & 336              \\ \midrule
RoBERTa-L \cite{liu2019roberta}     &     \begin{tabular}[c]{@{}c@{}}BooksCorpus and \\ English Wikipedia\end{tabular}       &               Masked Modeling  \cite{liu2019roberta}        & 355   \\ \midrule
GPT-2  \cite{radford2019language} &       WebText        &      Autoregression   & 774     \\ \midrule
text-ada-002     &          -  &            -        & - \\ 
\bottomrule
\end{tabular}%
}
\vspace{-0.25in}
\end{table}

\textbf{Setup}. For vision models, we use ResNet152 \cite{he2015resnet} with dimensions widened by a factor of two (ResNet152x2) fully supervised pre-trained on ImageNet-21K \cite{kolesnikov2020big}, Swin-L \cite{liu2021swin} fully supervised pre-trained on ImageNet-21K, EfficientNet-B3 semi-supervised pre-trained on noisy JFT-300M \cite{hinton2015distilling,chollet2017xception} and ImageNet-1K, and ViT-L \cite{dosovitskiy2020image} and ConvNext-L \cite{liu2022convnext} contrastive pre-trained on noisy Laion-2B \cite{cherti2023reproducible}. 
The details of the pre-trained models are shown in \cref{tab:vision-model-details} and all of them are adapted from TIMM \cite{rw2019timm}.
We evaluate the models on the 14 downstream ID and 4 OOD vision datasets.
We mainly compare our method with MLP tuning and LP, where we fine-tuning the modules using AdamW \cite{kingma2014adam} for $30$ epochs with a cosine learning rate scheduler. 
We set the learning rate as $0.1$ and weight decay of $0$ for LP, and set the learning rate as $0.001$ and weight decay of $1$e$-4$ for MLP tuning and black-box NMTune.

\begin{table}[t!]
\centering
\caption{Results on popular vision models that are pre-trained on noisy datasets. We report the avarage accuracy on 14 in-domain (ID) and 4 out-of-domain (OOD) tasks. }
\vspace{-0.1in}
\resizebox{0.95 \linewidth}{!}{
\begin{tabular}{l|l|cc|cc}
\toprule
\multicolumn{1}{c|}{\multirow{2}{*}{\begin{tabular}[c]{@{}c@{}}Pre-trained \\ Model\end{tabular}}} &
  \multicolumn{1}{c|}{\multirow{2}{*}{\begin{tabular}[c]{@{}c@{}}Tuning \\ Method\end{tabular}}} &
  \multicolumn{2}{c|}{In-Domain} &
  \multicolumn{2}{c}{Out-of-Domain} \\ \cline{3-6} 
\multicolumn{1}{c|}{} &
  \multicolumn{1}{c|}{} &
  Acc. &
  F1 &
  Acc. &
  F1 \\ \midrule
\multirow{3}{*}{\begin{tabular}[c]{@{}l@{}}JFT300M\\ Semi-Supervised \\ EfficientNet-B3 \end{tabular}} &
  LP &
  76.72 &
  0.3815 &
  44.13 &
  0.3594 \\
 &
  MLP &
  76.87 &
  0.3833 &
  45.95 &
  0.3624 \\
 &
  Ours &
  \textbf{77.63} &
  \textbf{0.3874} &
   \textbf{46.84} &
   \textbf{0.3654} \\ \midrule
\multirow{3}{*}{\begin{tabular}[c]{@{}l@{}}ImageNet-21K\\ Fully Supervised \\ ResNetv2-152x2 \end{tabular}} &
  LP &
  77.51 &
  0.3718 &
  40.82 &
  0.3062 \\
 &
  MLP &
  77.58 &
  0.3726 &
  41.73 &
  0.3053 \\
 &
  Ours &
  \textbf{78.43} &
  \textbf{0.3862} &
  \textbf{42.42} &
  \textbf{0.3100} \\ \midrule
\multirow{3}{*}{\begin{tabular}[c]{@{}l@{}}ImageNet-21K\\ Fully Supervised \\ Swin-L \end{tabular}} &
  LP &
  81.91 &
  0.4092 &
  50.88 &
  0.3838 \\
 &
  MLP &
  82.51 &
  0.4128 &
  51.21 &
  0.3811 \\
 &
  Ours &
  \textbf{84.16} &
  \textbf{0.4177} &
  \textbf{52.35} &
  \textbf{0.3901} \\ \midrule
\multirow{3}{*}{\begin{tabular}[c]{@{}l@{}}Laion-2B\\ CLIP \\ ConvNext-L \end{tabular}} &
  LP &
  88.86 &
  0.4432 &
  66.86 &
  0.4253 \\
 &
  MLP &
  88.53 &
  0.4417 &
  68.43 &
  0.4304 \\
 &
  Ours &
  \textbf{89.48} &
  \textbf{0.4457} &
  \textbf{70.30} &
   \textbf{0.4367} \\ \midrule
\multirow{3}{*}{\begin{tabular}[c]{@{}l@{}}Laion-2B\\ CLIP \\ ViT-L \end{tabular}} &
  LP &
  86.85 &
  0.4328 &
  66.89 &
  0.4208 \\
 &
  MLP &
  87.23 &
  0.4375 &
  69.50 &
  0.4221 \\
 &
  Ours &
  \textbf{88.57} &
  \textbf{0.4414} &
  \textbf{70.47} &
  \textbf{0.4246} \\ 
\bottomrule
\end{tabular}%
}%
\label{tab:res-vision}
\vspace{-0.15in}
\end{table}

\textbf{Results}. 
We present the average accuracy and F1 score across different datasets with three runs on vision models in \cref{tab:res-vision}.
Our method improves the quality of the noisy pre-trained features with better accuracy and F1 score on both ID and OOD vision tasks. 
A large margin on downstream tasks across different pre-training architectures and datasets is present by NMTune, demonstrating better features is learned. 
Noteworthy is that, although the MLP tuning also improves the performance in general, its performance gain is much smaller compared to our method, showing the effectiveness of the proposed method on mitigating the malicious effect of noise and improving generalization.


\subsection{Language Models and Datasets}

\textbf{Setup}. 
We evaluate BERT-L \cite{devlin2018bert}, RoBERTa-L \cite{liu2019roberta}, and GPT-2 \cite{radford2019language} on the GLUE \cite{2018glue} and GLUE-X~\cite{yang2022glue} benchmarks for ID and OOD performance.
The details of each model and are shown in \cref{tab:vision-model-details}. 
BERT-L and RoBERTa-L are pre-trained on the combination of the BooksCorpus data \cite{zhu2015bookscorpus} and English Wikipedia with uncompressed raw text. 
It is found that the raw pre-training data of BERT can be reduced from 16GB to 12GB with data cleaning \cite{yang2019xlnet}. 
GPT-2 is pre-trained on WebText \cite{radford2019language}, a scraped web dataset from Common Crawl that contains low-quality raw texts.
We also leverage OpenAI's API service ``text-ada-002''. 
We cannot use larger and more recent language models such as LLaMA \cite{touvron2023llama}, since they are unable to fit in a single V100 GPU and we are unsure whether GLUE is in their training data.
We use the AdamW optimizer and set the learning rate for LP as $0.01$ and for others as $0.001$ for all the experiments of language models. 
For LP, we do not use weight decay, and for others we use a weight decay of $0.0001$.
All tuning methods are trained for $10$ epochs with a linear scheduler.


\begin{table}[t!]
\centering
\caption{Evaluation on language models in practice that are pre-trained on noisy datasets. We use GLUE for in-domain (ID) tasks and GLUE-X for out-of-domain (OOD) tasks.}
\vspace{-0.1in}
\resizebox{0.85 \linewidth}{!}{%
\begin{tabular}{@{}l|l|cc@{}}
\toprule
\multicolumn{1}{c|}{\multirow{2}{*}{\begin{tabular}[c]{@{}c@{}}Pre-trained\\ Model\end{tabular}}} &
  \multicolumn{1}{c|}{\multirow{2}{*}{\begin{tabular}[c]{@{}c@{}}Tuning\\ Method\end{tabular}}} &
  \multirow{2}{*}{In-Domain} &
  \multirow{2}{*}{Out-of-Domain} \\
\multicolumn{1}{c|}{}      & \multicolumn{1}{c|}{} &  &  \\ \midrule
\multirow{3}{*}{BERT-L }    & LP                    & 69.44 & 50.65 \\
                           & MLP                   & 69.78 & 50.62 \\
                           & Ours                  & \textbf{70.26} & \textbf{51.63} \\ \midrule
\multirow{3}{*}{RoBERTa-L } & LP                    &  69.75 & 44.55 \\
                           & MLP                   &  70.27 & 45.22 \\
                           & Ours                  &  \textbf{70.97} &  \textbf{47.01} \\ \midrule
\multirow{3}{*}{GPT-2 }     & LP                    &  58.67 & 36.68 \\
                           & MLP                   &  58.44 & 37.24 \\
                           & Ours                  &  \textbf{59.34} &  \textbf{39.07} \\  \midrule
\multirow{3}{*}{text-ada-002}     & LP               & 56.96 & 44.06 \\
                           & MLP                   & 63.89 & 51.30  \\
                           & Ours                  & \textbf{65.99} &  \textbf{53.48} \\ 
\bottomrule
\end{tabular}%
}
\label{tab:nlp-results}
\vspace{-0.15in}
\end{table}

\textbf{Results}. 
In \cref{tab:nlp-results}, \method consistently achieves the best generalization performance.
It presents superior performance gain, especially on OOD tasks of GLUE-X. 
On the ``text-ada-002'' model with only API access, it also outperforms LP significantly, demonstrating the necessity of mitigating the effect of noise for better generalization. 
Interestingly, on the ID tasks of GLUE, we also observe a smaller gap of MLP tuning method to LP even with more parameters, showing that the MLP alone may not mitigate the influence of noisy data in pre-training. 

\begin{figure}[!t]
\centering
    \subfloat[ResNet-50 ID]{\label{fig:r50_cifar100_id_results}\includegraphics[width=0.48\linewidth]{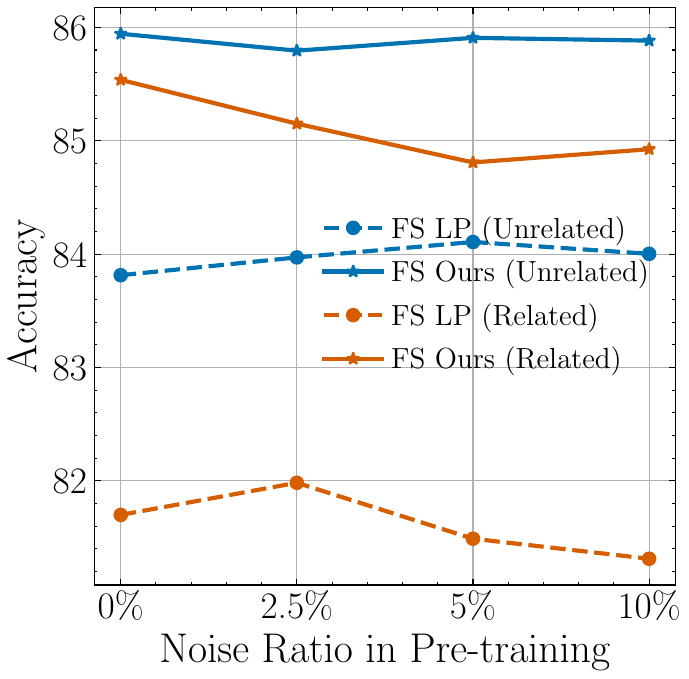}}
    \hfill
    \subfloat[ResNet-50 OOD]{\label{fig:r50_cifar100_ood_results}\includegraphics[width=0.48\linewidth]{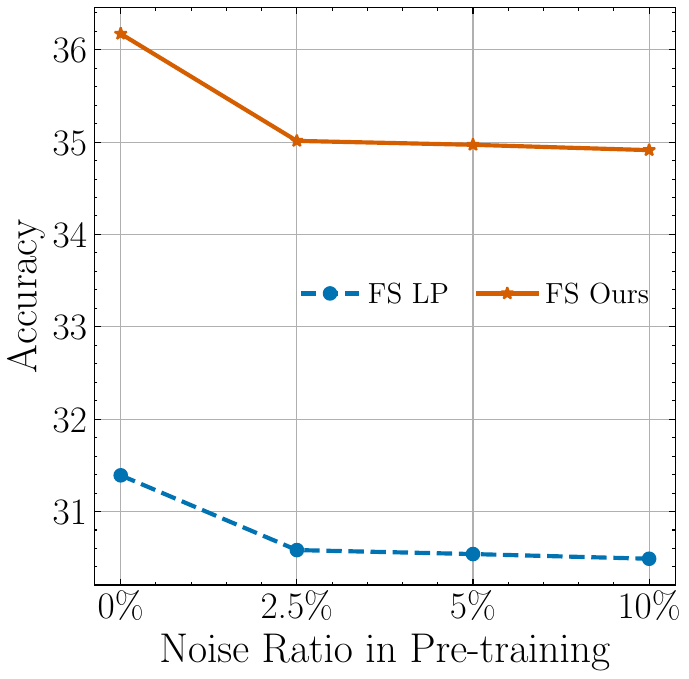}}
\caption{
Evaluation of IN-1K FS pre-trained ResNet-50 with asymmetric noise using our method (NMTune) in a black-box manner on ID and OOD downstream tasks, compared to LP. (a) Average accuracy on ID tasks, divided to tasks related to the asymmetric noise and unrelated to the noise; (b) Average accuracy on OOD tasks. 
With asymmetric noise, slight noise still benefits ID performance, even on noise related tasks.
Our method presents better performance with smaller difference between the clean and noisy pre-trained models.} 
\label{fig:r50_cifar100_results}
\vspace{-0.2in}
\end{figure}

\textbf{Results}.
For ID tasks, we report the average accuracy for noise related and noise unrelated tasks, as shown in \cref{fig:r50_cifar100_id_results}. 
One can still observe that the slight asymmetric pre-training noise can benefit LP ID performance for unrelated tasks. 
However, compared to unrelated tasks, the beneficial noise ratio decreases in related tasks, and performance degrades earlier. 
For OOD tasks, asymmetric noise consistently hurts performance, which aligns with our previous observations again.
Moreover, black-box NMTune can also facilitate downstream performance and rectify the malicious behaviors of models. 

\begin{figure*}[!t]
\centering
    \hfill
    \subfloat[IN-1K CIFAR10]{\label{fig:append-noise-lp-in1k-c10}\includegraphics[width=0.24\linewidth]{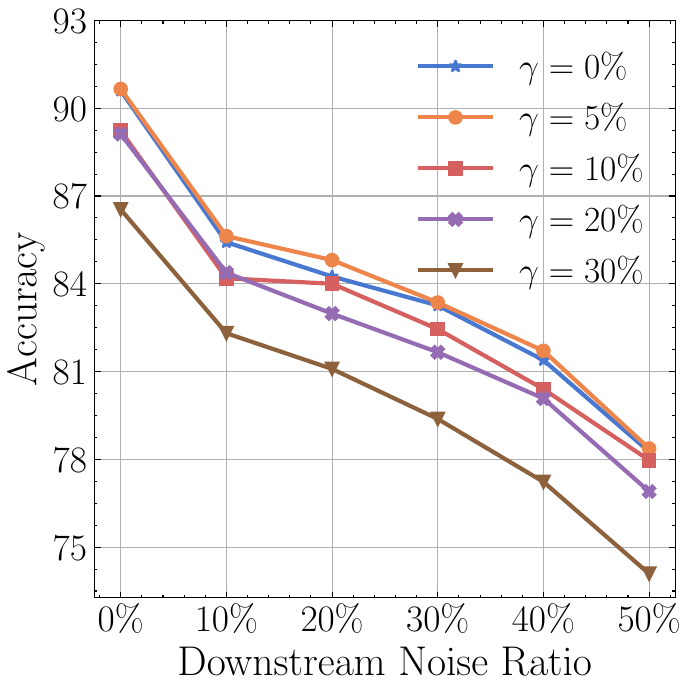}}
    \hfill
    \subfloat[IN-1K CIFAR100]{\label{fig:append-noise-lp-in1k-c100}\includegraphics[width=0.24\linewidth]{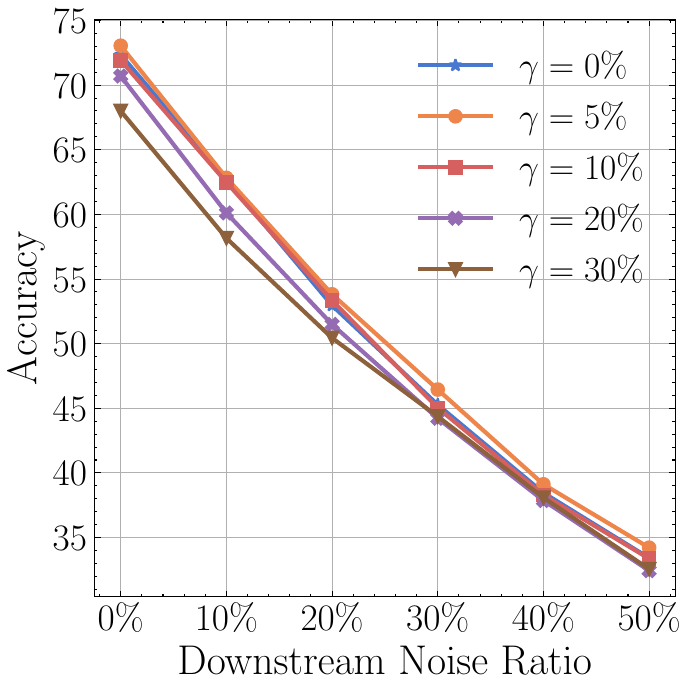}}
    \hfill
    \subfloat[YFCC15M CIFAR10]{\label{fig:append-noise-lp-yfcc-c10}\includegraphics[width=0.24\linewidth]{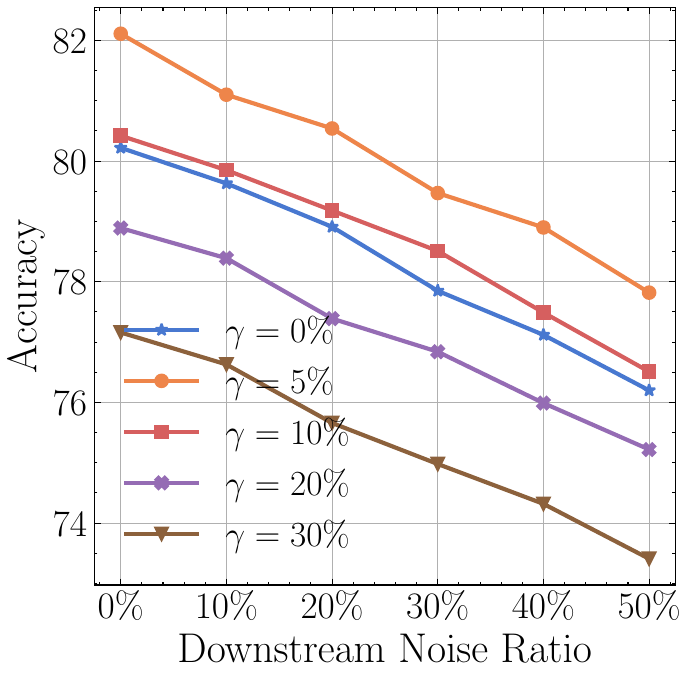}}
    \hfill
    \subfloat[YFCC15M CIFAR100]{\label{fig:append-noise-lp-yfcc-c100}\includegraphics[width=0.24\linewidth]{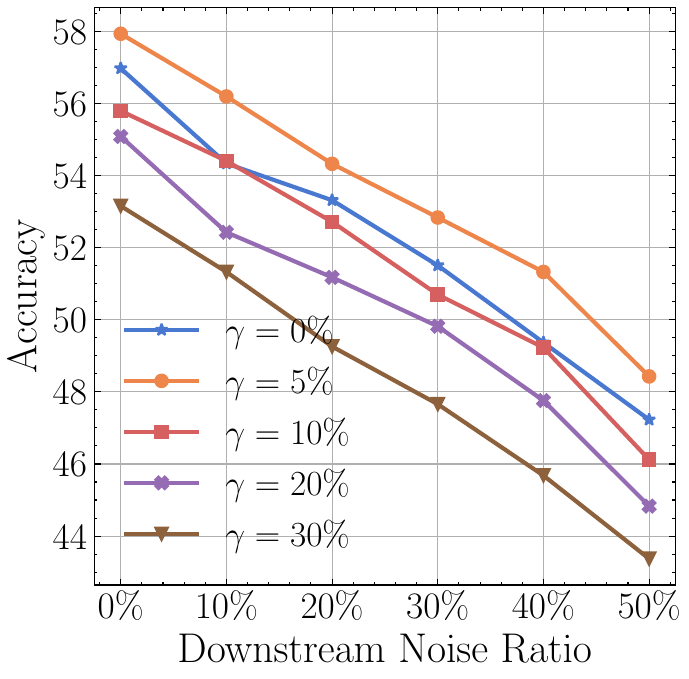}}
    \hfill

    \hfill
    \subfloat[IN-1K CIFAR10]{\label{fig:append-noise-nml-in1k-c10}\includegraphics[width=0.24\linewidth]{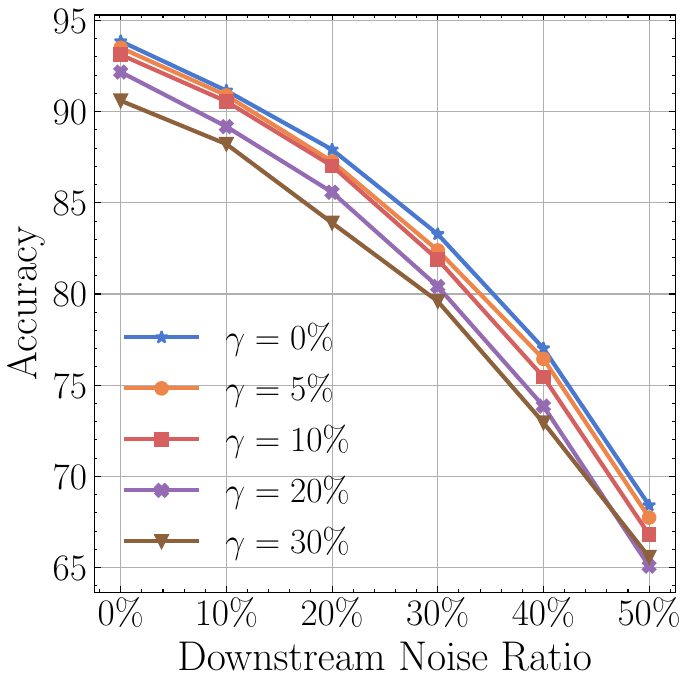}}
    \hfill
    \subfloat[IN-1K CIFAR100]{\label{fig:append-noise-nml-in1k-c100}\includegraphics[width=0.24\linewidth]{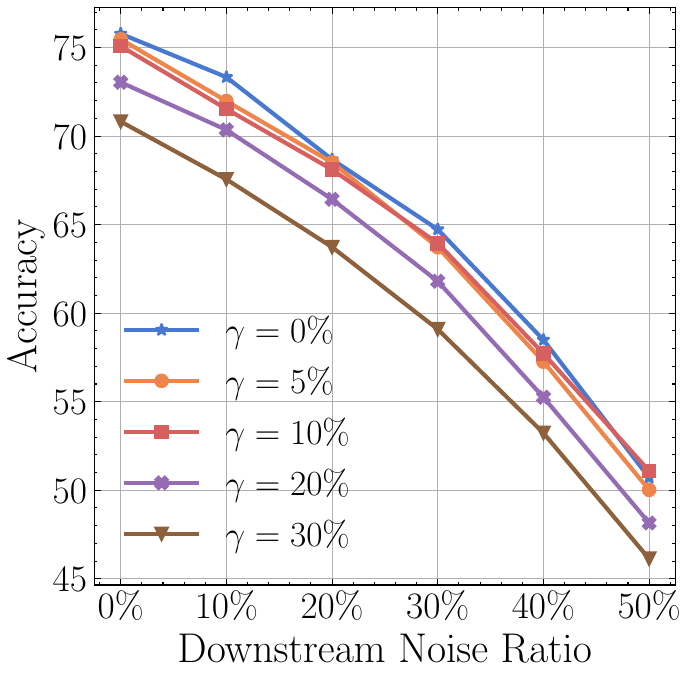}}
    \hfill
    \subfloat[YFCC15M CIFAR10]{\label{fig:append-noise-nml-yfcc-c10}\includegraphics[width=0.24\linewidth]{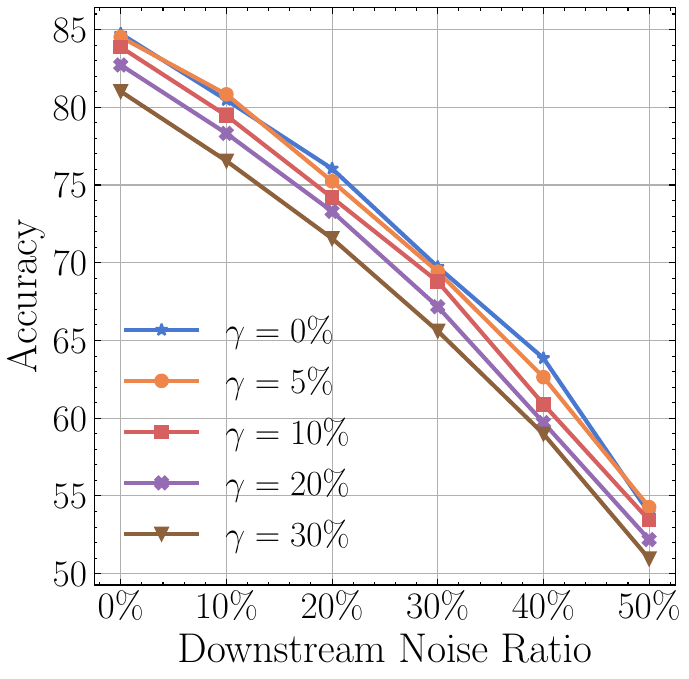}}
    \hfill
    \subfloat[YFCC15M CIFAR100]{\label{fig:append-noise-nml-yfcc-c100}\includegraphics[width=0.24\linewidth]{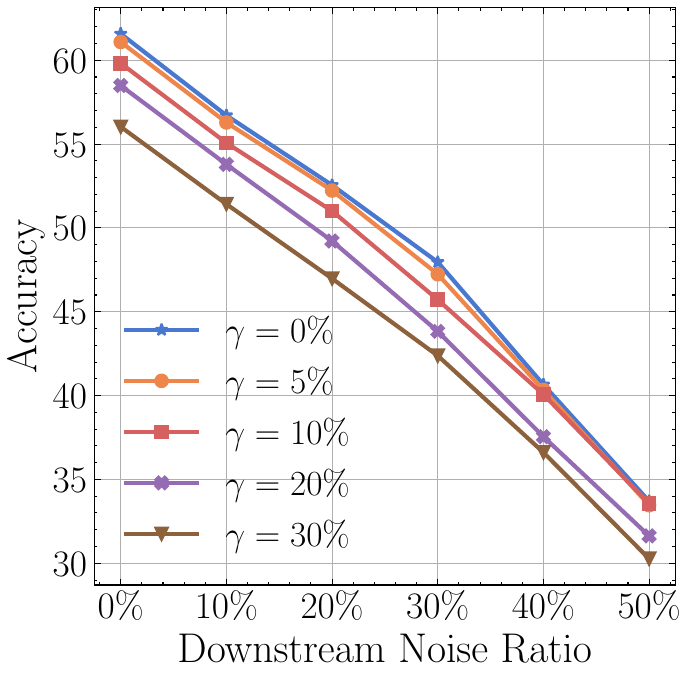}}
    \hfill
\caption{Linear Probing and NMTune of noisy ResNet-50 models on noisy CIFAR-10 and CIFAR-100.} 
\label{fig:append-noise-lp}
\vspace{-0.2in}
\end{figure*}

\subsection{Asymmetric Noise in Pre-training}
\label{sec:asym-noise}

\textbf{Setup}.
In previous experiments and analyses, we mainly studied with random pre-training noise, where the concept/label of the supervision can be corrupted by any other concepts/labels.
Here, we additionally verify with asymmetric and non-fully-random pre-training noise for IN-1K FS training of ResNet-50, where the noise only exists in a restricted subset of concepts/labels. 
To introduce such noise in IN-1K FS pre-training, we select the classes in CIFAR-100 as the subset of concepts/labels.
Given the class labels of CIFAR-100, we use NLTK \cite{bird2009natural} to find their WordNet synonyms in the class labels of IN-1K.  
We then introduce noise to only the image samples related to these synonyms with noise ratios of $\{0\%, 2.5\%, 5\%, 10\%\}$. 
We use linear probing and black-box NMTune to evaluate the effect of asymmetric noise in pre-training on ID classification tasks of CIFAR-10, CIFAR-100, Food101, Caltech101, and EuroSAT, and OOD tasks of DomainNet, following the same hyper-parameters in \cref{sec:understand}.
For these downstream tasks, CIFAR-10, CIFAR-100, and DomainNet are directly related to the asymmetric pre-training noise, while others are not. 

\subsection{Noisy model \revision{transfer} learning with noisy label learning}
\label{sec:noisy-label}

\textbf{Setup}.
We additionally study the setting where both pre-training and downstream datasets contain noisy labels. 
For pre-training noise, we use the ResNet-50 models FS pre-trained on noisy IN-1K and CLIP pre-trained on noisy YFCC15M with different noise ratios $\gamma \in \{0\%, 5\%, 10\%, 20\%, 30\%\}$, as in \cref{sec:understand}. 
For downstream noise, we adopt synthetic noise CIFAR-10 and CIFAR-100 which are usually used in noisy label learning \cite{sopliu22w,chen2023imprecise}. 
We generate label noise by uniformly flipping labels for a percentage of the downstream training set.
We denote the noise ratio at downstream as $\eta \in \{0\%, 10\%, 20\%, 30\%, 40\%, 50\%\}$. 
We compare LP and black-box NMTune and adopt the same hyper-parameters.

\textbf{Results}.
The results are shown in \cref{fig:append-noise-lp}.
On the LP results, we find similar observations as our analysis in \cref{sec:understand}, where the $5\%$ and $10\%$ noisy pre-trained models usually outperform the clean pre-trained model on downstream tasks, even the tasks contain different levels of downstream noise. 
It indicates that the effects of pre-training noise also extend to noisy downstream tasks, which highlights the importance of the proposed new topic - Noisy Model \revision{Transfer} Learning - as complementary to noisy label learning. 
More importantly, we find that the proposed NMTune method has a similar mitigation effect on noisy downstream tasks as clean ones. 
The NMTune results show that the clean pre-trained models now produce superior performance compared to noisy pre-trained models. 
It also improves the general performance when the noise ratio in downstream tasks is light, e.g., smaller than $40\%$. 
When the noise ratio in downstream tasks further increases, the performance of NMTune falls short of LP, which is acceptable because the regularization terms are not designed to be noise-tolerant. 
It is noteworthy that, even with a slightly worse performance than LP, the performance of clean pre-trained mode still stays the best with NMTune. 

\subsection{Running time analysis}
Here, we present the average GPU hours of NMTune, MLP tuning, and LP.
Runtime analysis for NMTune, compared to LP and MLP tuning, is shown in \cref{tab:append-exp-runtime}.  
All of our downstream experiments are performed on a single NVIDIA V100 GPU.
Thus, we report the average GPU hours for running the ID and OOD evaluation of vision and language tasks. 
From the results, the proposed NMTune introduces minimal computation, compared to MLP with exactly the same parameters.
The additional computation burden may involve the SVD calculation and the covariance matrix calculation on the features.

\begin{table}[h]
\centering
\caption{Average run time for LP, MLP, and NMTune in terms of GPU hours across in-domain and out-of-domain vision and language datasets.}
\label{tab:append-exp-runtime}
\vspace{-0.1in}
\resizebox{0.4 \textwidth}{!}{%
\begin{tabular}{@{}l|lll@{}}
\toprule
Datasets                    & LP   & MLP  & Ours \\ \midrule
Vision In-Domain (14)       & 5.01 & 7.96 & 8.23 \\
Vision Out-of-Domain (4)    & 2.19 & 4.48 & 4.56 \\ \midrule
Language In-Domain (8)      & 2.87 & 3.76 & 3.82 \\
Language Out-of-Domain (14) & 1.21 & 1.34 & 1.45 \\ \bottomrule
\end{tabular}%
}
\vspace{-0.2in}
\end{table}

\section{Conclusion and Limitation}
We presented \textit{Noisy Model \revision{Transfer} Learning}, a new research direction for understanding and mitigating the effect of label noise in pre-training on downstream tasks. 
Extensive experiments demonstrate that proper noise in pre-training can benefit in-domain tasks and hurt out-of-domain tasks. 
We then proposed NMTune to mitigate the malignant effect of noise and improve the generalization performance of various noisy pre-trained models and APIs. 
We stand at the threshold of a new topic in machine learning research, where the nuanced understanding of noise's role could unlock unprecedented levels of model robustness and adaptability. 
It is our fervent hope that our initial foray into "Noisy Model Learning" will ignite and inspire more efforts, pushing the boundaries of understanding of large foundation models.

This research highlights two primary areas where further exploration could yield significant advancements.
First, our investigation into the effects of noise has been confined to supervised pre-training, despite the prevalent use of both supervised and self-supervised learning frameworks in the field.
Supervised learning focuses on learning a direct correlation between inputs and their corresponding labels \cite{he2015resnet,radford2021learning}, whereas self-supervised learning does not rely on labels, but predicts parts of the data itself, such as masked modeling \cite{devlin2018bert,he2022mae}. 
Although our analysis focuses on the supervised setting, it can be extended to self-supervised learning that can also be viewed as supervised learning on the data itself.
Second, while the largest pre-training data scale we studied is around 27M, it is still relatively small compared to the pre-training data scale of production foundation models. 
However, with experiments on data scales ranging from 1M to 27M, we demonstrate that our observations manifest themselves on all scales and possibly also on larger scales, which we have yet to explore. 


%



\bibliographystyle{IEEEtran}
\bibliography{IEEEabrv,ref}







\begin{IEEEbiography}[{
\includegraphics[width=1in,height=1.25in,clip,keepaspectratio]{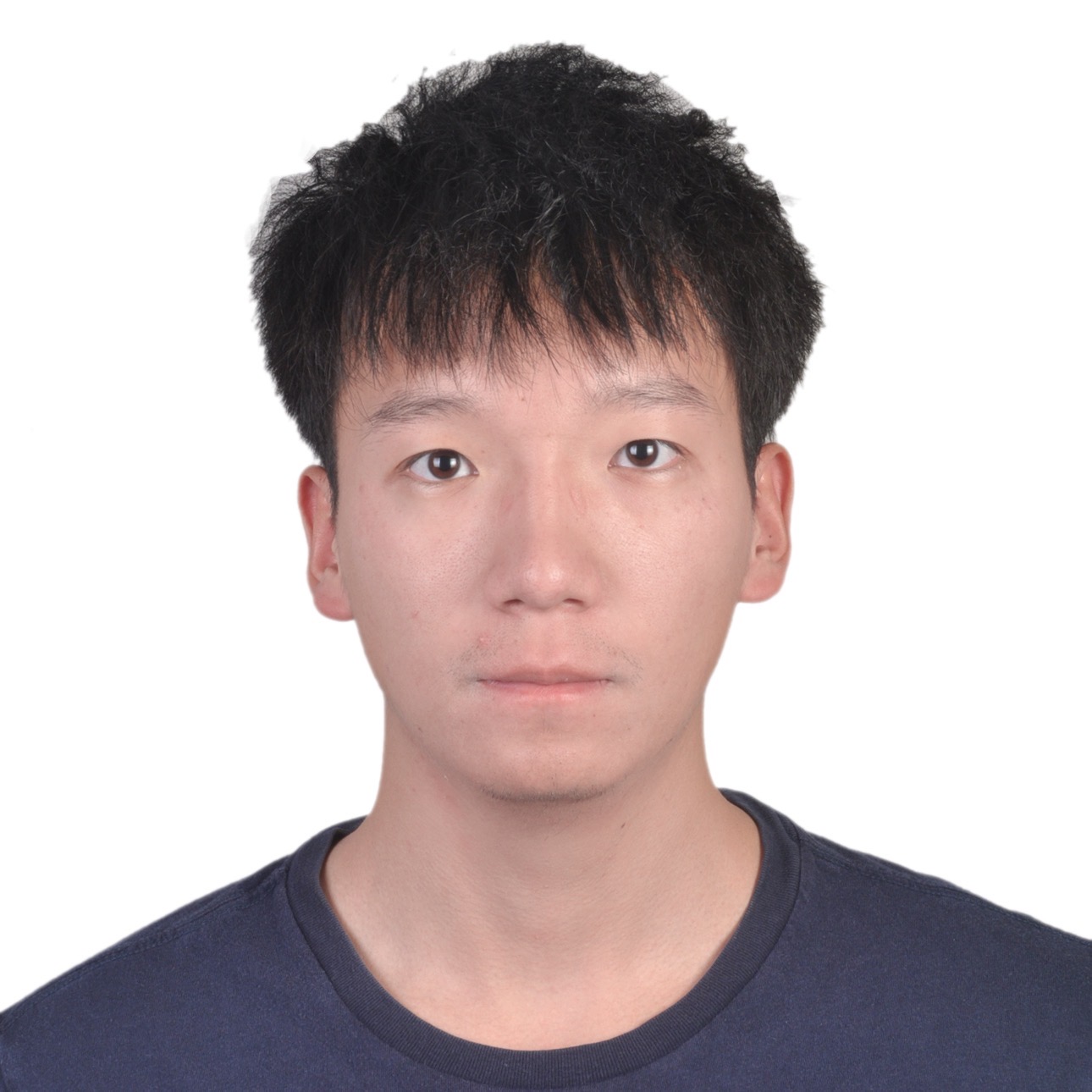}}]{Hao Chen} is currently a Ph.D. candidate at Carnegie Mellon University, advised by Prof. Bhiksha Raj. He received a M.S. degree in Electrical and Computer Engineering from Carnegie Mellon University and a B.S. degree jointly from the University of Edinburgh and Tianjin University, respectively. 
His research interests lie in robustness and generalization of foundation models and weak supervision.
\end{IEEEbiography}

\begin{IEEEbiography}[{
\includegraphics[width=1in,height=1.25in,clip,keepaspectratio]{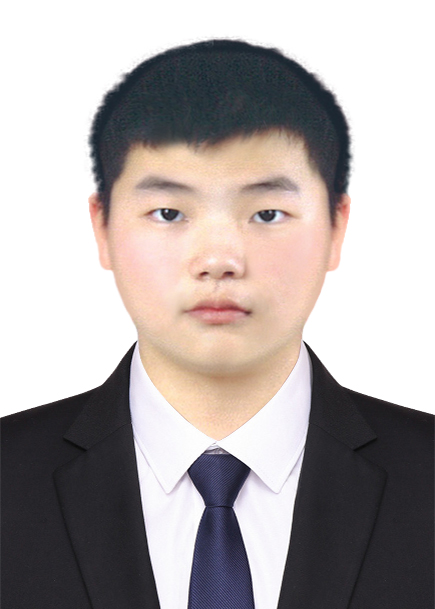}}]{Zihan Wang} is currently a Master of Computer Vision(MSCV) Student in Robotics Institute at Carnegie Mellon University, advised by Prof.Deva Ramanan. He received a B.S. degree from the University of Edinburgh. His research interests lie in machine learning, computer vision and robotics.
\end{IEEEbiography}

\begin{IEEEbiography}[{
\includegraphics[width=1in,height=1.25in,clip,keepaspectratio]{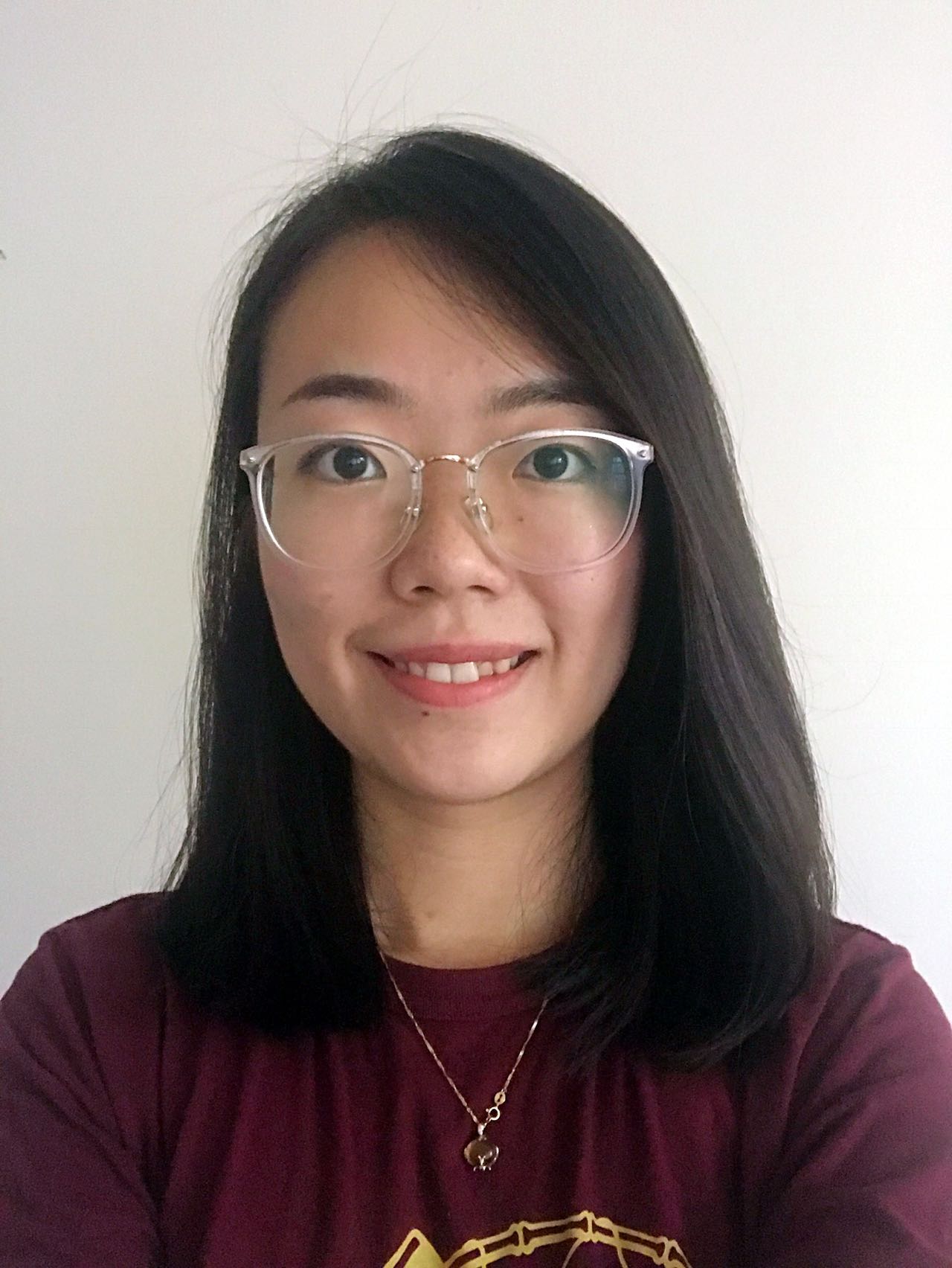}}]{Ran Tao} is currently a machine learning technical lead at Metalenz, Boston, US. She received her Ph.D. degree in Electrical and Computer Engineering in 2023 from Carnegie Mellon University and a M.S degree in Machine Learning from Carnegie Mellon University. Her research interests lie in weakly-supervised learning and its applications on computer vision tasks. 
\end{IEEEbiography}

\begin{IEEEbiography}[{
\includegraphics[width=1in,height=1.25in,clip,keepaspectratio]{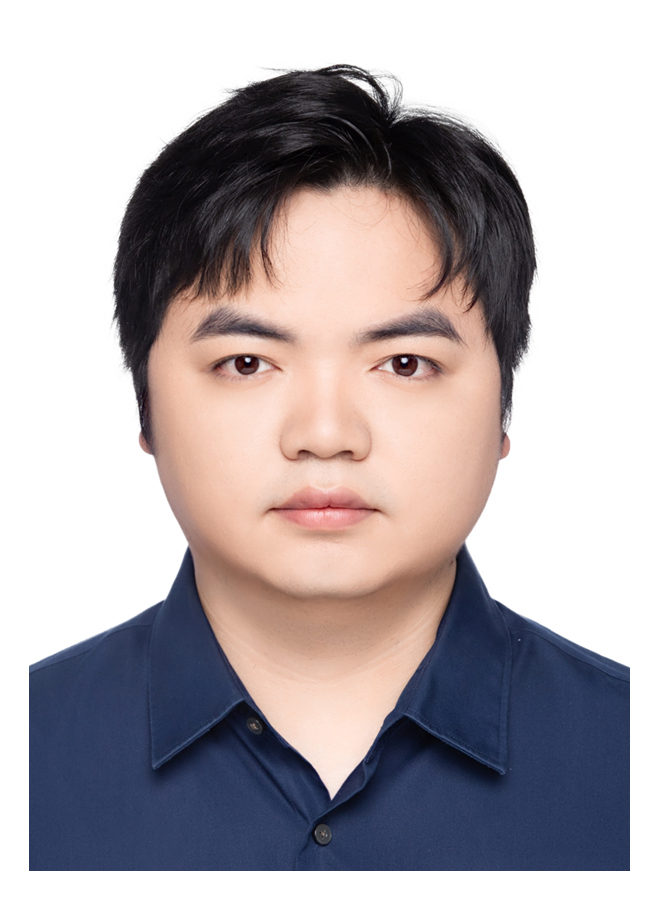}}]{Hongxin Wei} is currently an Assistant Professor in the Department of Statistics and Data Science at Southern University of Science and Technology, Shenzhen, China. He received his Ph.D. degree at the School of Computer Science and Engineering, Nanyang Technological University, in 2023. He was a visiting scholar at the University of Wisconsin Madison in 2022. His research interests mainly include data-centric machine learning and foundation model. He is the reviewer or PC member of several leading journals and conferences such as TPAMI, ICLR, ICML, NeurIPS, etc.
\end{IEEEbiography}

\begin{IEEEbiography}[{\includegraphics[width=1in,height=1.25in,clip,keepaspectratio]{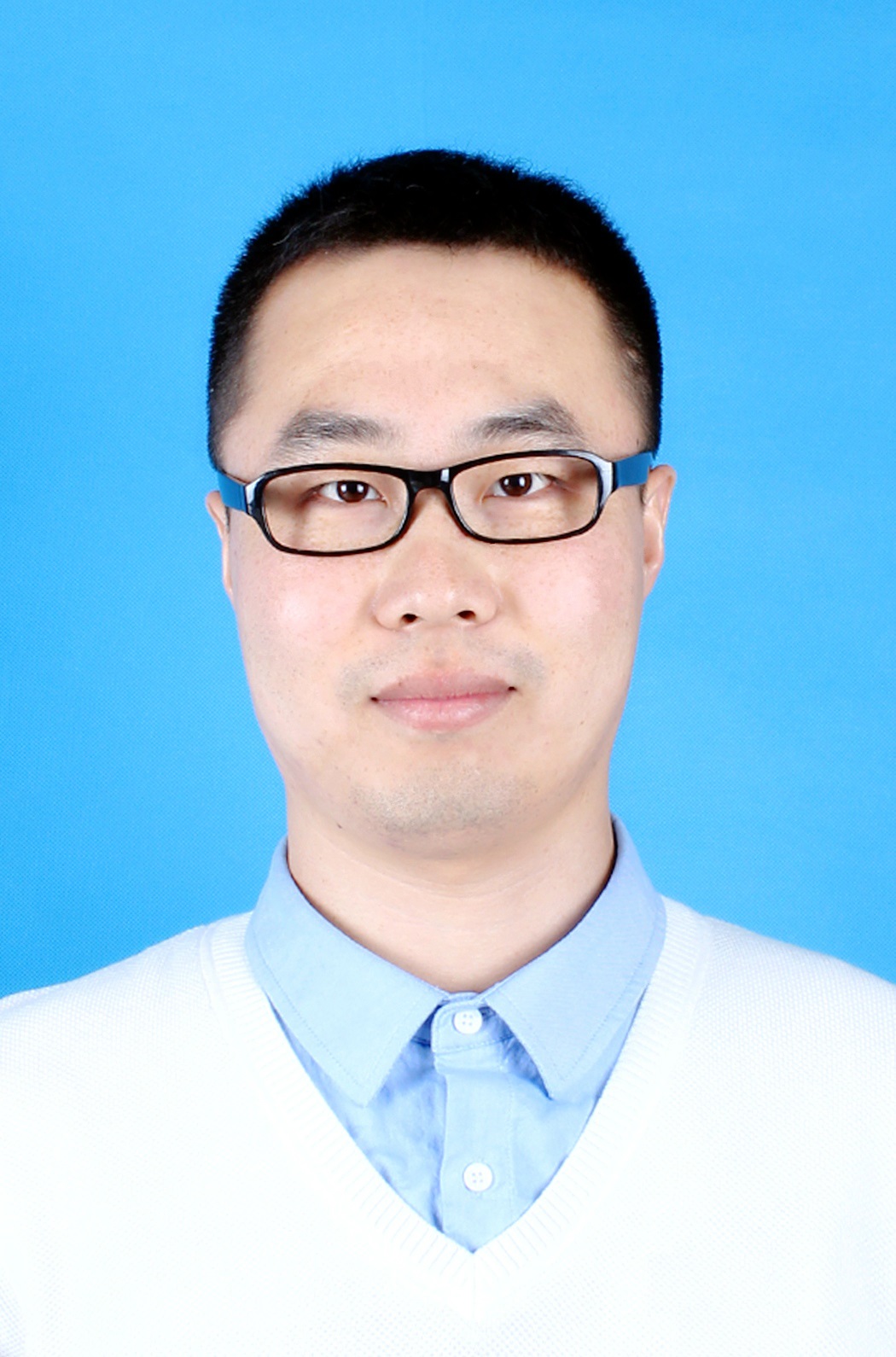}}]{Jindong Wang} is currently an assistant professor at William \& Mary.
He was a Senior Researcher at Microsoft Research Asia, Beijing, China. 
He received his Ph.D. degree from the Institute of Computing Technology, Chinese Academy of Sciences, Beijing, China, in 2019. 
He was a visiting student at Hong Kong University of Science and Technology (HKUST) in 2018. 
His research interest mainly includes transfer learning, machine learning, data mining, and ubiquitous computing. 
He serves as the publicity co-chair of IJCAI’19 and session chair of ICDM’19. He is the reviewer or PC member of several leading journals and conferences such as TPAMI, TKDE, ICLR, ICML, NeurIPS, CVPR, etc.
\end{IEEEbiography}

\begin{IEEEbiography}[{\includegraphics[width=1in,height=1.25in,clip,keepaspectratio]{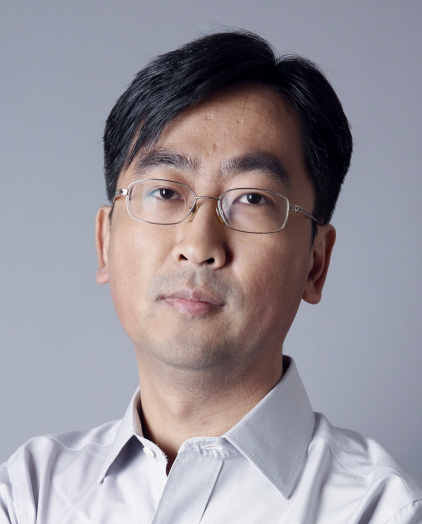}}]{Xing Xie} 
is currently a senior principal research manager at Microsoft Research Asia. 
He is also a distinguished member of ACM, an IEEE Fellow, and a fellow of China Computer Federation (CCF).
His research interest mainly includes data mining, social computing and ubiquitous computing.
During the past years, he has published over 300 referred journal and conference papers, won the 10-year impact award in ACM SIGSPATIAL 2019, the best student paper award in KDD 2016, and the best paper awards in ICDM 2013 and UIC 2010. 
He currently serves on the editorial boards of ACM TSC, ACM TIST, ACM IMWUT, GeoInformatica, PMC, and CCF TPCI. 
In recent years, he was involved in the program or organizing committees of over 70 conferences and workshops. 
Especially, he served as program co-chair of ACM Ubicomp 2011, PCC 2012, UIC 2015, SMP 2017, and ACM SIGSPATIAL 2020. 
\end{IEEEbiography}

\begin{IEEEbiography}[{
\includegraphics[width=1in,height=1.25in,clip,keepaspectratio]{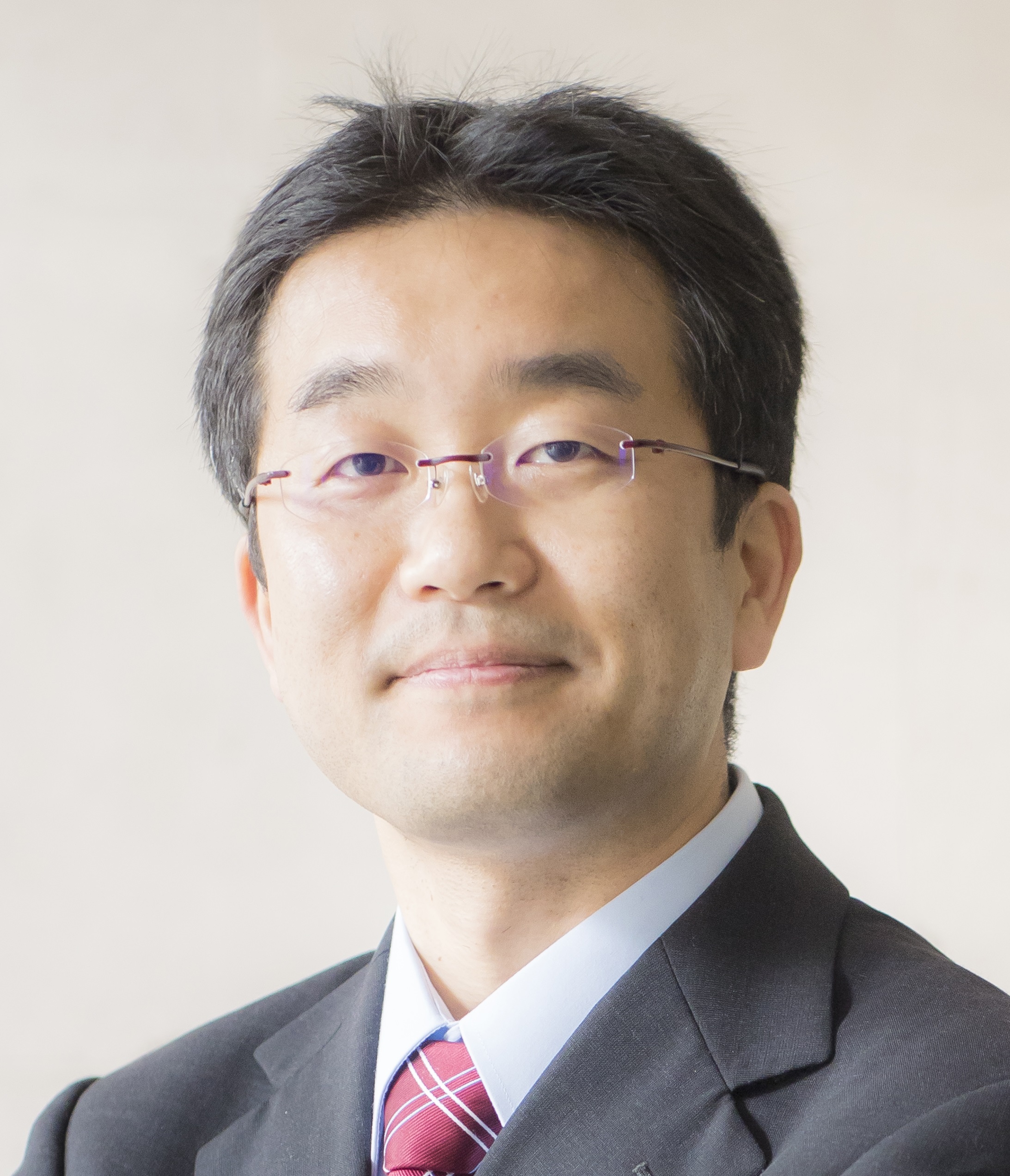}}]{Masashi Sugiyama} received a Ph.D. degree in Computer Science from Tokyo Institute of Technology, Japan, in 2001. After experiencing
assistant and associate professors at the same institute, he became a professor at the University of Tokyo in 2014. Since 2016, he has concurrently served as Director of the RIKEN Center for Advanced Intelligence Project. His research interests include theories and algorithms of machine learning. He was a recipient of the Japan Academy Medal in 2017 and the Commendation for Science and Technology by the Minister of Education, Culture, Sports, Science and Technology in Japan in 2022.
\end{IEEEbiography}

\begin{IEEEbiography}[{
\includegraphics[width=1in,height=1.25in,clip,keepaspectratio]{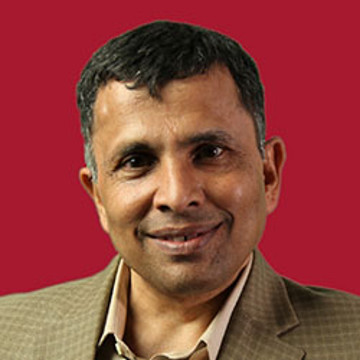}}]{Bhiksha Raj} received the PhD degree in Electrical and Computer Engineering from Carnegie Mellon University, Pittsburgh, PA, USA, in 2000. He is currently a professor with Computer Science Department, Carnegie Mellon University where he leads the Machine Learning for Signal Processing Group. He joined the Carnegie Mellon faculty in 2009, after spending time with the Compaq Cambridge Research Labs and Mitsubishi Electric Research Labs. He has devoted his career to developing speech- and audio-processing technology. He has had several seminal contributions in the areas of robust speech recognition, audio content analysis and signal enhancement, and has pioneered the area of privacy-preserving speech processing. He is also the chief architect of the popular Sphinx-4 speech-recognition system.
\end{IEEEbiography}

\end{document}